\documentclass[10pt,twocolumn,letterpaper]{article}

\usepackage{iccv}
\usepackage{times}
\usepackage{epsfig}
\usepackage{graphicx}
\usepackage[cmex10]{amsmath}
\usepackage{amsthm}
\usepackage{amssymb}
\usepackage{tabulary,color}
\usepackage{tabularx}
\usepackage{multirow}
\usepackage{booktabs, siunitx, dcolumn}
\usepackage{colortbl}
\usepackage{pifont}
\usepackage{dsfont}
\usepackage{enumitem} 
\usepackage[para]{threeparttable} 
\usepackage[export]{adjustbox}
\usepackage[percent]{overpic}
\usepackage{array} 
\usepackage{url} 
\usepackage{fixltx2e} 
\usepackage{algorithm}     
\usepackage{algorithmic}     
\usepackage[caption=false,font=footnotesize]{subfig}

\usepackage{pdfpages}

\usepackage[pagebackref=true,breaklinks=true,letterpaper=true,colorlinks,bookmarks=false]{hyperref}

\iccvfinalcopy 

\ificcvfinal\fi
\begin{document}

\title{Deformable kernel networks for guided depth map upsampling}

\author{Beomjun Kim\\
Yonsei University\\
\and
Jean Ponce\\
Inria~/~PSL Research University\\
\and
Bumsub Ham\\ 
Yonsei University\\
}

\maketitle

\vspace{-0.5cm}
\begin{abstract}
\vspace{-0.3cm}
We address the problem of upsampling a low-resolution (LR) depth map using a registered high-resolution (HR) color image of the same scene. Previous methods based on convolutional neural networks~(CNNs) combine nonlinear activations of spatially-invariant kernels to estimate structural details from LR depth and HR color images, and regress upsampling results directly from the networks. In this paper, we revisit the weighted averaging process that has been widely used to transfer structural details from hand-crafted visual features to LR depth maps. We instead learn explicitly sparse and spatially-variant kernels for this task. To this end, we propose a CNN architecture and its efficient implementation, called the deformable kernel network~(DKN), that outputs sparse sets of neighbors and the corresponding weights adaptively for each pixel. We also propose a fast version of DKN~(FDKN) that runs about $17$ times faster~($0.01$ seconds for a HR image of size $640 \times 480$). Experimental results on standard benchmarks demonstrate the effectiveness of our approach. In particular, we show that the weighted averaging process with $3 \times 3$ kernels~(i.e.,~aggregating $9$ samples sparsely chosen) outperforms the state of the art by a significant margin.
\end{abstract}

\begin{figure}[t]
\centering
\includegraphics[width=0.94\linewidth]{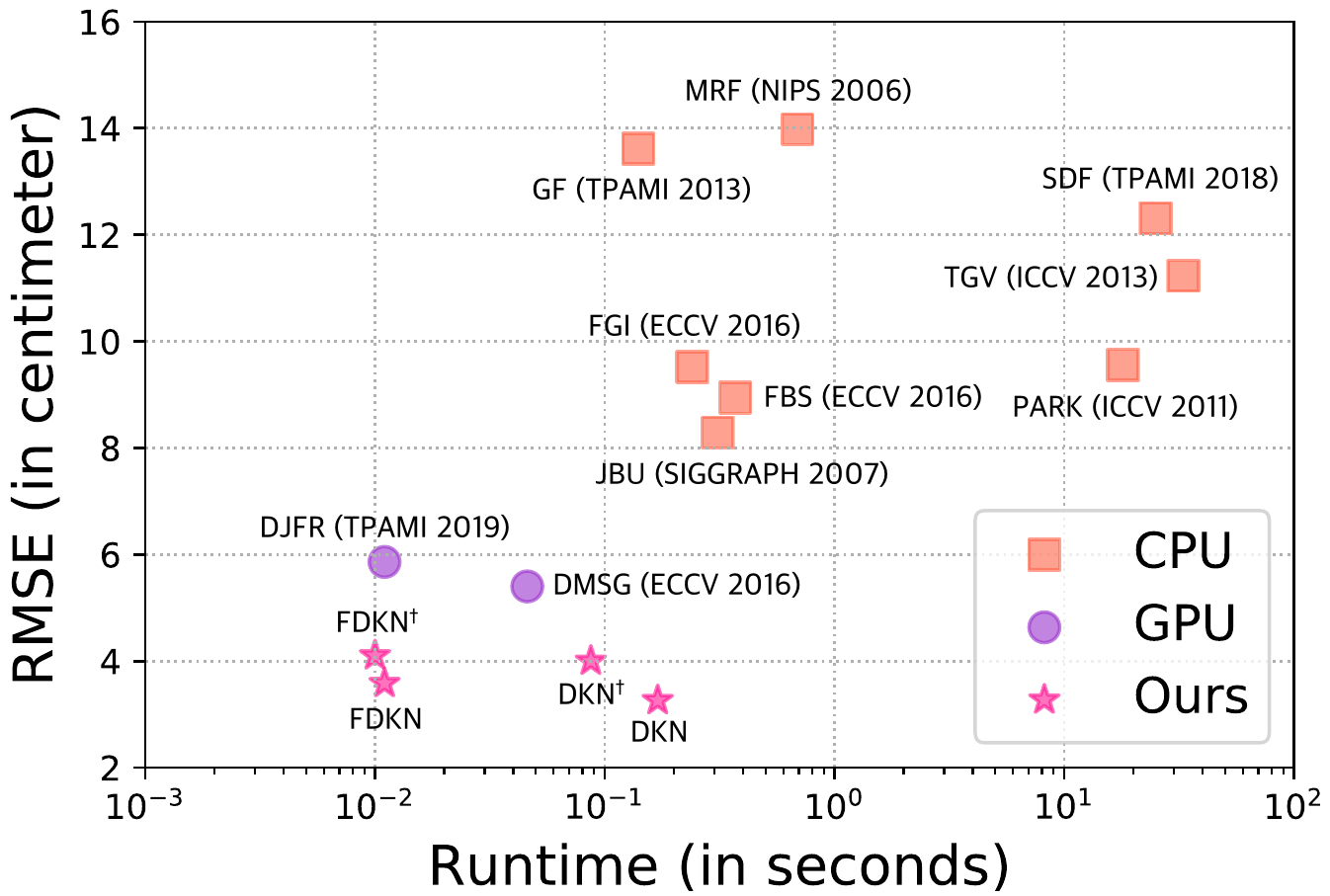}
\caption{Runtime and root mean squared errors~(RMSE) comparison of upsampled depth maps~($8 \times$) on the NYU v2~\cite{silberman2012indoor} dataset. We decrease the error by $30 \%$ to $50 \%$ on average compared to the state of the art, which is quite significant, with essentially zero loss in speed. $\dagger$: Our models trained with the LR depth map only without any guidance. See Secs.~\ref{sec:exp} or~\ref{sec:overview} for details.}
\label{fig:timegraph}
\vspace{-0.2cm}
\end{figure}

\vspace{-0.2cm}
\section{Introduction}\label{sec:introduction}
\vspace{-0.2cm}
Acquiring depth information is one of the fundamental tasks in computer vision, for scene recognition~\cite{hoffman2016learning}, pose estimation~\cite{shotton2011real} and 3D reconstruction~\cite{furukawa2010accurate}, for example. Recent stereo matching methods based on convolutional neural networks~(CNNs)~\cite{luo2016efficient,zbontar2015computing} give high-quality depth maps, but still require a huge computational cost, especially in the case of a large search range. Consumer depth cameras~(\eg,~the ASUS Xtion Pro~\cite{asus2018xtion} and the Microsoft Kinect~\cite{zhang2012kinect}), typically coupled with RGB sensors, are practical alternatives to obtain depth maps at low cost. Although they provide dense depth maps, these typically offer limited spatial resolution and depth accuracy. To address this problem, registered high-resolution~(HR) color images can be used as guidance to enhance the spatial resolution of low-resolution~(LR) depth maps~\cite{ferstl2013,Gu2017learning,ham2018robust,hui2016depth,kopf2007joint,li2016deep,li2016fast,park2011,yang07}. The basic idea behind this approach, called guided or joint image filtering, is to exploit their statistical correlation to transfer structural details from the guidance HR color image to the target LR depth maps, typically by estimating spatially-variant kernels from the guidance. Concretely, given the target image $f$ and the guidance image $g$, the filtering output $\hat f$ at position ${\bf{p}}=(x,y)$ is expressed as a weighted average~\cite{he2013guided,kopf2007joint,tomasi1998bilateral,wu2018fast}:
\vspace{-0.1cm}
\begin{equation}\label{eq:weighted_average}
\hat f_{\bf{p}} = \sum_{{\bf{q}} \in \mathcal{N({\bf{p}})}} W_{{\bf{p}}{\bf{q}}}(g)f_{\bf{q}},
\vspace{-0.1cm}
\end{equation}
where we denote by $\mathcal{N(\bf{p})}$ a set of neighbors (defined on a discrete regular grid) near the position $\bf{p}$. The filter kernel $W$ is typically a function of the guidance image $g$~\cite{ferstl2013,he2013guided,kopf2007joint,park2011}, normalized so that 
\vspace{-0.1cm}
\begin{equation}\label{eq:constraint}
	\sum_{{\bf{q}} \in \mathcal{N({\bf{p}})}} W_{{\bf{p}}{\bf{q}}}(g) = 1.
\vspace{-0.1cm}
\end{equation}

Classical approaches to depth map upsampling mainly focus on designing the filter kernels $W$ and the set of neighbors $\mathcal{N}$~(\ie,~sampling locations $\bf{q}$). They use hand-crafted kernels and predefined neighbors without learning~\cite{he2013guided,kopf2007joint,tomasi1998bilateral}. For example, the guided filter~\cite{he2013guided} uses spatially-variant matting Laplacian kernels~\cite{levin2008closed} to encode local structures from the HR color image. These methods use regularly sampled neighbors for aggregating pixels, and do not handle inconsistent structures in the HR color and LR depth images, causing texture-copying artifacts~\cite{ferstl2013}. To address the problem, both HR color and LR depth images have been used to extract common structures~\cite{Gu2017learning,ham2018robust,li2016deep,li2017joint}. Recently, learning-based approaches using CNNs~\cite{hui2016depth,li2016deep,li2017joint} have also become increasingly popular. The networks are trained using large quantities of data, capturing natural image priors and often outperforming traditional methods by large margins. These methods do not use a weighted averaging process. They combine instead nonlinear activations of spatially-invariant kernels learned from the networks. That is, they approximate spatially-variant kernels by mixing the activations of spatially-invariant ones nonlinearly~(\eg, via the ReLU function~\cite{krizhevsky2012imagenet}). 

 In this paper, we propose to exploit spatially-variant kernels explicitly to encode the structural details from both HR color and LR depth images as in classical approaches, but learn the kernel weights in a supervised manner. We also learn the set of neighbors, building an adaptive and sparse neighborhood system for each pixel. This also allows sub-pixel information aggregation, which may be difficult to achieve by hand. To implement this idea, we propose a CNN architecture and its efficient implementation, called a \emph{deformable kernel network}~(DKN), for learning sampling locations of the neighboring pixels and their corresponding kernel weights at every pixel. We also propose a fast version of DKN~(FDKN), achieving a $17$ times speed-up compared to the plain DKN for a HR image of size $640 \times 480$, while retaining its superior performance. We show that the weighted averaging process, even trained with the LR depth map only without any guidance~(\ie,~$g=f$ in~\eqref{eq:weighted_average}), with $9$ points sparsely sampled, is sufficient to obtain a new state of the art~(Fig.~\ref{fig:timegraph}). Our code and models are available online: \url{https://cvlab-yonsei.github.io/projects/DKN}

\noindent {\textbf{Contributions.}} The main contributions of this paper can be summarized as follows:
\begin{itemize}[leftmargin=*]
\vspace{-0.2cm}
  \item[$\bullet$] We introduce a novel variant of the classical guided weighted averaging process for depth map upsapling and its implementation, the DKN, that computes the set of neighbors and their corresponding weights adaptively for individual pixels.
\vspace{-0.2cm}
  \item[$\bullet$] We propose a fast version of DKN~(FDKN) that runs about $17$ times faster than the DKN while retaining its superior performance.
\vspace{-0.2cm}
  \item[$\bullet$] We achieve a new state of the art, outperforming all existing methods we are aware of by a large margin, and clearly demonstrating the advantage of our approach to learning both kernel weights and sampling locations. We also provide an extensive experimental analysis to investigate the influence of all the components and parameters of our model. 
\vspace{-0.2cm}
\end{itemize}

\vspace{-0.2cm}
\section{Related work}
\vspace{-0.2cm}
Here we briefly describe the context of our approach, and review representative works related to ours. 

\begin{figure*}
\centering
\includegraphics[width=0.9\linewidth]{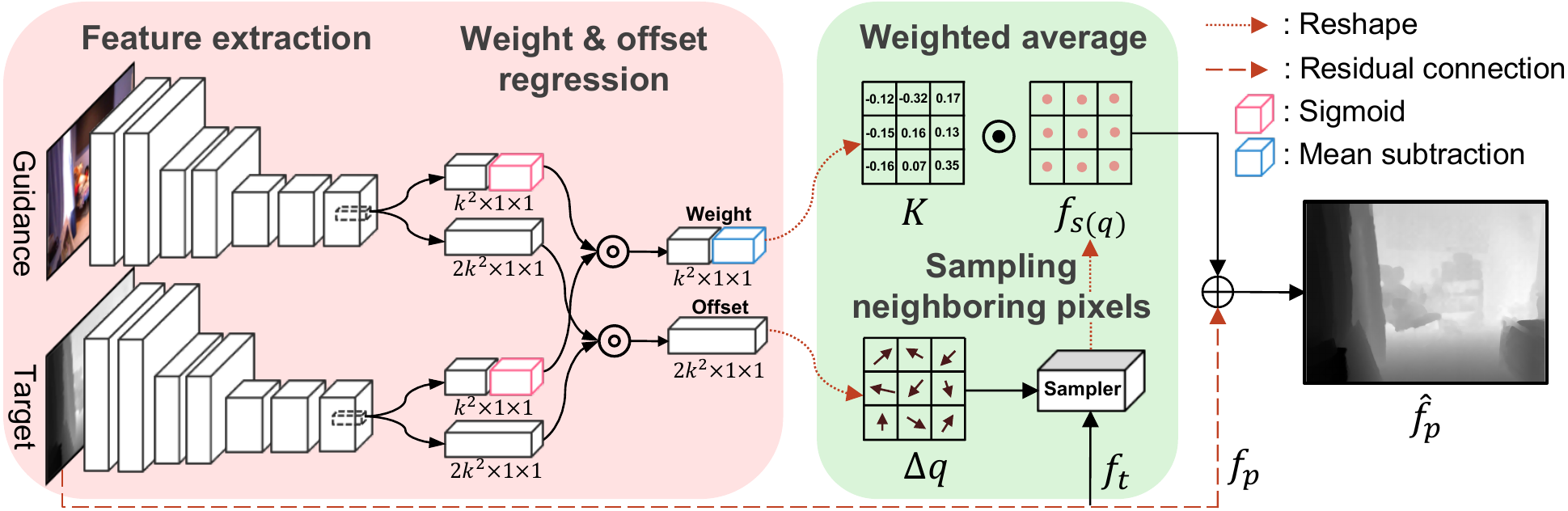}
\caption{The DKN architecture. We learn the kernel weights~$K$ and the spatial sampling offsets~$\Delta {\bf{q}}$ from the feature maps of HR color and LR depth images. To obtain the residual image~$\hat f_{\bf{p}} -f_{\bf{p}}$, we then compute the weighted average with the kernel weights~$K$ and image values~$f_{{\bf{s}}({\bf{q}})}$ sampled at offset locations~$\Delta {\bf{q}}$ from the neighbors~$f_{{\bf{t}}}$. Finally, the result is combined with the LR depth image~$f_{{\bf{p}}}$ to obtain the upsampling result~$\hat f_{\bf{p}}$. Our model is fully convolutional and is learned end-to-end. We denote by $\circledcirc$ and $\odot$~element-wise multiplication and dot product, respectively. The reshaping operator and residual connection are drawn in dotted and dashed lines, respectively. See Table~\ref{table:architecture} for the detailed description of the network structure.~(Best viewed in color.) }
\label{fig:overview}
\vspace{-0.2cm}
\end{figure*}

\noindent \textbf{Depth map upsampling.}
We categorize depth map upsampling into explicit/implicit weighted-average methods and learning-based ones. First, explicit weighted-average methods compute the output at each pixel by a weighted average of neighboring pixels in the LR depth image, where the weights are estimated from the HR color image~\cite{he2013guided,kopf2007joint} to transfer fine-grained structures. The bilateral~\cite{kopf2007joint,tomasi1998bilateral} and guided~\cite{he2013guided} filters are representative methods that have been successfully adapted to depth map upsampling. They use hand-crafted kernels to estimate the weights, which may transfer erroneous structures to the target image~\cite{li2016deep}. Second, implicit weighted-average methods formulate depth map upsampling as an optimization problem, and minimize an objective function that usually involves fidelity and regularization terms~\cite{ferstl2013,ham2018robust,Gu2017learning,li2016fast,park2011}. The fidelity term encourages the output to be close to the LR depth image, and the regularization term gives the output having a structure similar to that of the HR color image. Although, unlike explicit ones, implicit weighted-average methods exploit global structures in the HR color image, hand-crafted regularizers may not capture structural priors. Finally, learning-based methods can further be categorized into dictionary-~and CNN-based approaches. Dictionary-based methods exploit the relationship between paired LR and HR depth patches, additionally coupled with the HR color image~\cite{ferstl2015variational,kwon2015data,yang2010image}. In CNN-based methods~\cite{hui2016depth,li2016deep,li2017joint}, an encoder-decoder architecture is typically used to learn features from the HR color and/or LR depth images, and the output is then regressed directly from the network. Other methods~\cite{riegler16gdsr,riegler16dsr} integrate a variational optimization into CNNs by unrolling the optimization steps of the primal-dual algorithm, which requires two stages in training and a number of iterations in testing. Similar to implicit weighted-average methods, they use hand-crafted regularizers, which may not capture structural priors.

Our method borrows from both explicit weighted-average methods and CNN-based ones. Unlike existing explicit weighted-average methods~\cite{he2013guided,kopf2007joint}, that use hand-crafted kernels and neighbors defined on a fixed regular grid, we leverage CNNs to learn the set of sparsely chosen neighbors and their corresponding weights adaptively. Our method differs from previous CNN-based ones~\cite{hui2016depth,li2016deep,li2017joint} in that we learn sparse and spatially-variant kernels for each pixel to obtain upsampling results as a weighted average. The bucketing stretch in single image super-resolution~\cite{getreuer2018blade,romano2017raisr} can be seen as a non-learning-based approach to filter selection. It assigns a single filter by solving a least-squares problem for a set of similar patches (buckets). In contrast, our model \emph{learns} different filters using CNNs even for similar RGB patches, since we learn them from a set of multi-modal images (\ie.,~pairs of RGB/D images). 

\noindent \textbf{Variants of the spatial transformer~\cite{jaderberg2015spatial}.}
Recent works introduce more flexible and effective CNN architectures. Jaderberg~\emph{et al}. propose a novel learnable module, the spatial transformer~\cite{jaderberg2015spatial}, that outputs the parameters of the desired spatial transformation~(\eg, affine or thin plate spline)~given a feature map or an input image. The spatial transformer makes a standard CNN for classification invariant to a set of geometric transformation, but it has a limited capability of handling local transformations. Most similar to ours are the dynamic filter network~\cite{jia2016dynamic} and its variants~(the adaptive convolution network~\cite{niklaus2017video} and the kernel prediction networks~\cite{bako2017kernel,mildenhall2018burst,vogels2018denoising}), where a set of local transformation parameters is generated adaptively conditioned on the input image. The main differences between our model and these works are two-fold. First, our network is more general in that it is not limited to learning spatially-variant kernels, but it also learns the sampling locations of neighbors. This allows to aggregate sparse but highly related samples only, enabling an efficient implementation in terms of speed and memory and achieving state-of-the-art results even with aggregating 9 samples sparsely chosen. For comparison, the adaptive convolution and kernel prediction networks require lots of samples~(\eg,~$21 \times 21$ in~\cite{bako2017kernel,vogels2018denoising}, $41 \times 41$ in~\cite{niklaus2017video}, and $8 \times 5 \times 5$ in~\cite{mildenhall2018burst}). As will be seen in our experiments, learning sampling locations of neighbors clearly boosts the performance significantly compared to learning kernel weights only. Second, as other guided image filtering approaches~\cite{ham2018robust, he2013guided,kopf2007joint,li2016deep,li2017joint}, our model is easily adapted to other tasks such as saliency map upsampling, cross-modality image restoration, texture removal, and semantic segmentation. We focus here on depth upsampling but see the supplement for some examples. In contrast, the adaptive convolution network is specialized to video frame interpolation, and kernel prediction networks are applicable to denoising Monte Carlo renderings~\cite{bako2017kernel,vogels2018denoising} or burst denoising~\cite{mildenhall2018burst} only. Our work is also related to the deformable convolutional network~\cite{dai2017deformable}. The basic idea of deformable convolutions is to add  offsets to the sampling locations defined on a regular grid in standard CNNs. The deformable convolutional network samples features directly from learned offsets, but shares the same weights for different sets of offsets as in standard CNNs. In contrast, we use spatially-variant weights for each sampling location.

\begin{table*}[t]
\centering
\footnotesize
\caption{Network architecture details. ``BN" and ``Res." denote the batch normalization~\cite{ioffe2015batch} and residual connection, respectively. We denote by ``DownConv" convolution with stride 2. The inputs of our network are 3-channel HR color and 1-channel LR depth images~(denoted by $D$). For the model without the residual connection, we use an L1 normalization layer~(denoted by ``L1 norm.") instead of subtracting mean values for weight regression.}
\label{table:architecture}
\label{table:depth-upsampling}
\addtolength{\tabcolsep}{-3pt}
\renewcommand{\arraystretch}{0}
\newcolumntype{L}[1]{>{\raggedright\arraybackslash}p{#1}}
\newcolumntype{C}[1]{>{\centering\arraybackslash}p{#1}}
\newcolumntype{R}[1]{>{\raggedleft\arraybackslash}p{#1}}
\begin{tabular}{L{4.5cm} R{2.5cm} L{4.5cm} R{2.5cm}}
\midrule
\multicolumn{2}{c}{Feature extraction} & \multicolumn{2}{c}{Weight regression}\\
\cmidrule(lr){1-2}
\cmidrule(lr){3-4}
\multicolumn{1}{c}{Type} & \multicolumn{1}{c}{Output} & \multicolumn{1}{c}{Type} & \multicolumn{1}{c}{Output} \\ 
\cmidrule(lr){1-1}
\cmidrule(lr){2-2}
\cmidrule(lr){3-3}
\cmidrule(lr){4-4}
Input~(Receptive field)& $ D \times 51 \times 51$ & Conv($1\times1$) & $k^2 \times 1 \times 1$ \\
\addlinespace[0.5em]
Conv($7\times7$)-BN-ReLU & $32\times 45 \times 45$ & Sigmoid & $k^2 \times 1 \times 1$\\
\addlinespace[0.5em]
DownConv($2\times2$)-ReLu & $ 32 \times 22 \times 22$ & Mean subtraction or & \multirow{2}{*}{$k^2 \times 1 \times 1$} \\
\addlinespace[0.5em]
Conv($5\times5$)-BN-ReLU & $ 64 \times 18 \times 18$ & L1 norm. (w/o Res.) \\

\cmidrule(lr){3-4} 
DownConv($2\times2$)-ReLU & $ 64 \times 9 \times 9$ &\multicolumn{2}{c}{Offset regression} \\
\cmidrule(lr){3-4} 
Conv($5\times5$)-BN-ReLU & $ 128 \times 5 \times 5$ & \multicolumn{1}{c}{Type} & \multicolumn{1}{c}{Output} \\
\cmidrule(lr){3-3}
\cmidrule(lr){4-4} 
Conv($3\times3$)-ReLU & $ 128 \times 3 \times 3$ & \multirow{2}{*}{Conv($1\times1$)} & \multirow{2}{*}{$2k^2 \times 1 \times 1$} \\
\addlinespace[0.5em]
Conv($3\times3$)-ReLU & $ 128 \times 1 \times 1$ \\
\midrule
\end{tabular}
\vspace{-0.4cm}
\end{table*}

\vspace{-0.2cm}
\section{Approach}\label{sec:proposed}
\vspace{-0.2cm}
In this section, we briefly describe our approach, and present a concrete network architecture. We then describe a fast version of DKN.
\vspace{-0.2cm}
\subsection{Overview}\label{sec:overview}
\vspace{-0.2cm}
Our network mainly consists of two parts~(Fig.~\ref{fig:overview}): We first learn spatially-variant kernel weights and spatial sampling offsets w.r.t the regular grid. To this end, a two-stream CNN~\cite{simonyan2014two}, where each sub-network has the same structure~(but different parameters), uses the guidance~(HR color) and target~(LR depth) images to extract features that are used to estimate the kernel weights and the offsets. We then compute a weighted average using the learned kernel weights and sampling locations computed from the offsets to obtain a residual image. Finally, the upsampling result is obtained by combining the residual image with the LR depth map. Note that we can train DKN without the residual connection, by directly computing the upsampling result as a weighted average. Note also that we can train our model \emph{without} the guidance of the HR color image. In this case, we use a single-stream CNN to extract features from the LR depth map only in both training and testing. Our network is fully convolutional, does not require fixed-size input images, and it is trained end-to-end.

\noindent {\textbf{Weight and offset learning.}}~Dual supervisory information for the weights and offsets is typically not available. We learn instead these parameters by minimizing directly the discrepancy between the output of the network and a reference HR depth map. In particular, constraints on weight and offset regression~(sigmoid and mean subtraction layers in Fig. 2) specify how the kernel weights and offsets behave and guide the learning process. For weight regression, we apply a sigmoid layer that makes all elements larger than 0 and smaller than 1. We then subtract the mean value from the output of the sigmoid layer so that the regressed weights should be similar to high-pass filters with kernel weights adding to 0. For offset regression, we do not apply the sigmoid layer, since relative offsets~(for~$x$,~$y$ positions)~from locations on a regular grid can have negative values.

\noindent {\textbf{Residual connection.}}~The main reason behind using a residual connection is that the upsampling result is largely correlated with the LR depth map, and both share low-frequency content~\cite{he2016residual,kim2016accurate,li2017joint,Zhang2017}. Focussing on learning the residuals also accelerates training speed while achieving better performance~\cite{kim2016accurate}. Note that contrary to~\cite{he2016residual,kim2016accurate,li2017joint,Zhang2017}, we obtain the residuals by a weighted averaging process with the learned kernels, instead of regressing them directly from the network output. Empirically, the kernels learned with the residual connection have the same characteristics as the high-pass filters widely used to extract important structures~(\eg, object boundaries)~from images~(See the supplemental material). 

\vspace{-0.2cm}
\subsection{DKN architecture}  
\vspace{-0.2cm}
We design a fully convolutional network to learn the kernel weights and the sampling offsets for individual pixels. We show in Table~\ref{table:architecture} the detailed description of the network structure.

\noindent
\textbf{Feature extraction.}
We adapt an architecture similar to~\cite{niklaus2017video} for feature extraction. It consists of 7 convolutional layers, two of which use convolutions with multiple strides~(``DownConv" in Table~\ref{table:architecture}), that enlarge a receptive field size with a small number of network parameters to estimate. We input the HR color and LR depth images to each of the sub-networks, resulting in a feature map of size~$128 \times 1 \times 1$ for a receptive field of size $51 \times 51$. The LR depth map is initially upsampled using bicubic interpolation. We use the ReLU~\cite{krizhevsky2012imagenet} as an activation function. Batch normalization~\cite{ioffe2015batch} is used for speeding up training and regularization. 

\noindent
\textbf{Weight regression.}
For each sub-network, we add a $1 \times 1$ convolutional layer on top of the feature extraction layer. It gives a feature map of size~$k^2 \times 1 \times 1$, where $k$ is the size of the filter kernel, which is used to regress the kernel weights. To estimate the weights, we apply a sigmoid layer to each feature map of size~$k^2 \times 1 \times 1$, and then combine the outputs by element-wise multiplication~(see Fig.~\ref{fig:overview}). We could use a softmax layer as in~\cite{bako2017kernel,niklaus2017video,vogels2018denoising}, but empirically find that it does not perform as well as the sigmoid layer. The softmax function encourages the estimated kernel to have only a few non-zero elements, which is not appropriate for estimating the weights for sparsely sampled pixels. The estimated kernels should be similar to high-pass filters, with kernel weights adding to 0. To this end, we subtract the mean value from the combined output of size $k^2 \times 1 \times 1$. For our model without a residual connection, we apply instead L1 normalization to the output of size $k^2 \times 1 \times 1$. Since the sigmoid layer makes all elements in the combined output larger than 0, applying L1 normalization forces the kernel weights to add to 1 as in~\eqref{eq:constraint}.

\noindent
\textbf{Offset regression.}
Similar to the weight regression case, we add a $1 \times 1$ convolutional layer on top of the feature extraction layer. The resulting two feature maps of size~$2k^2 \times 1 \times 1$ are combined by element-wise multiplication. The final output contains relative offsets (for $x$, $y$ positions) from locations on a regular grid. In our implementation, we use $3\times 3$ kernels, but the output is computed by aggregating 9 samples sparsely chosen from a much larger neighborhood. The two main reasons behind the use of small-size kernels are as follows:~(1)~This enables an efficient implementation in terms of speed and memory. (2) The reliability of samples are more important than the total number of samples aggregated. As will be seen in Sec.~\ref{sec:exp}, our model outperforms the guided filter~\cite{he2013guided} using kernels of size $17 \times 17$ by a large margin. A similar finding is noted in~\cite{wang2007optimized}, which shows that only high-confidence samples should be chosen when estimating foreground and background images in image matting. Note that offset regression is closely related to nonlocal means~\cite{buades2005non} in that both select which pixels to aggregate instead of immediate neighbors. Likewise, learning offsets is related to ``self-supervised" correspondence models in stereo matching~\cite{godard2017unsupervised} and optical flow estimation~\cite{jason2016back}. For example, in the case of stereo matching, a model is trained to produce a flow field such that a right image is reconstructed by a left one according to that flow field. Our model with filter kernels of size $k\times k$ computes $k^2$ correspondences for each pixel within input images, and also learns the corresponding matching confidence~(\ie,~the kernel weights).
\begin{figure}[t]
  \centering
  \subfloat[]{
    \begin{minipage}[b]{0.27\linewidth}
      \includegraphics[width=\linewidth]{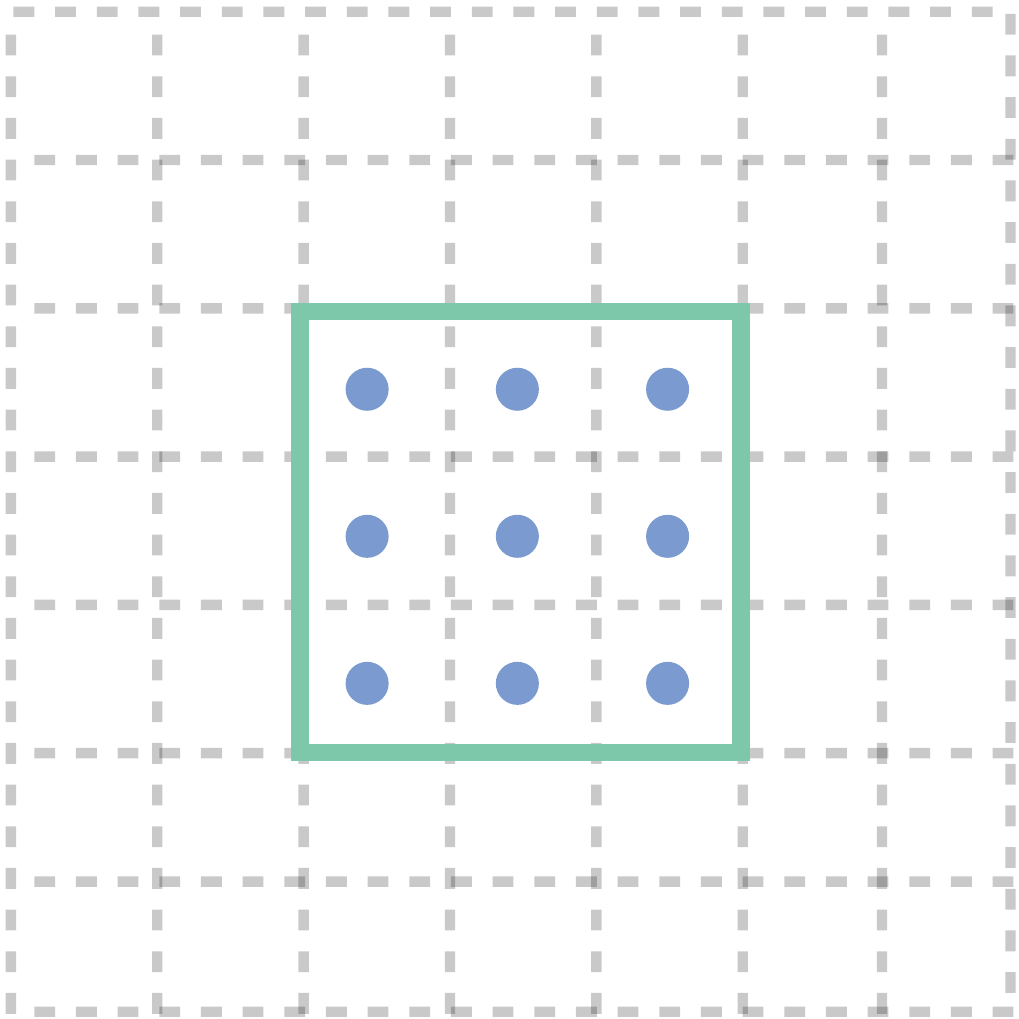} 
    \end{minipage}
  }
  \subfloat[]{
    \begin{minipage}[b]{0.27\linewidth}
      \includegraphics[width=\linewidth]{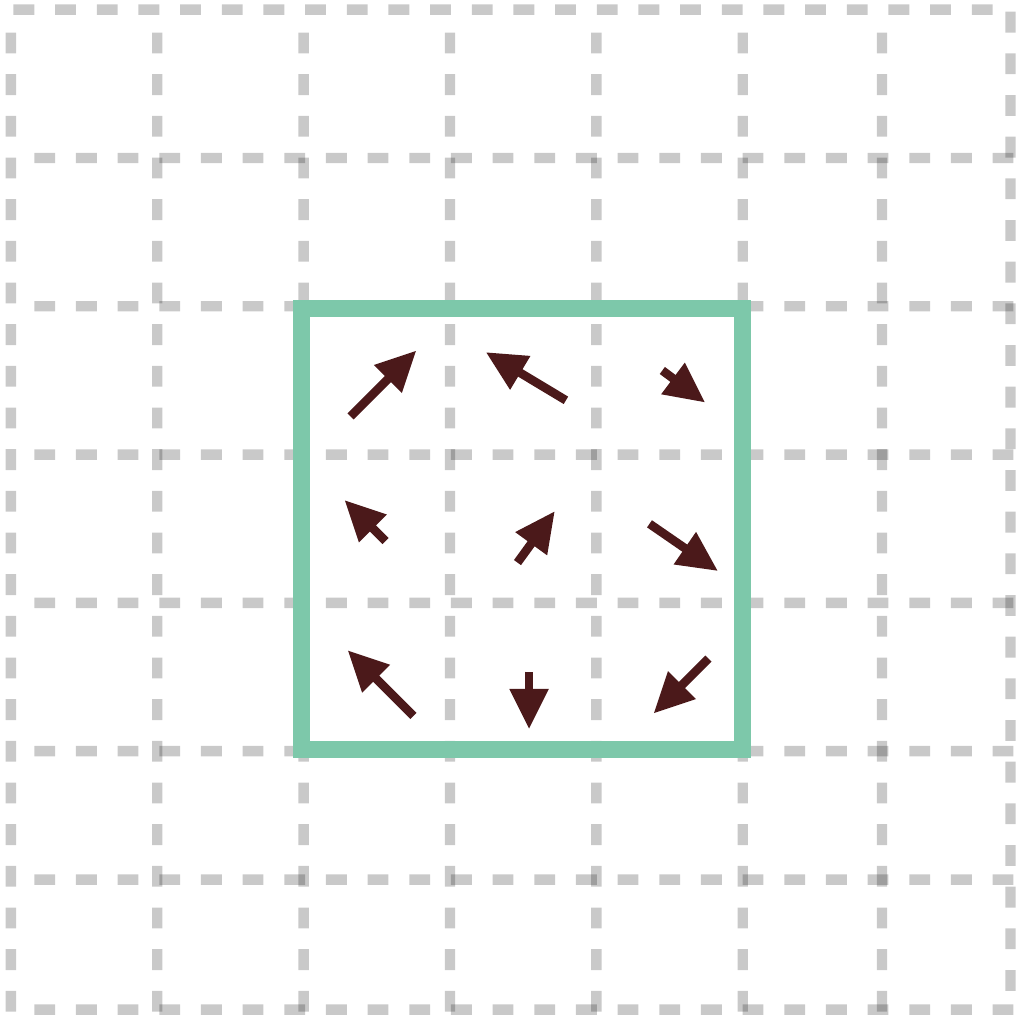} 
    \end{minipage}
  }
    \subfloat[]{
    \begin{minipage}[b]{0.27\linewidth}
      \includegraphics[width=\linewidth]{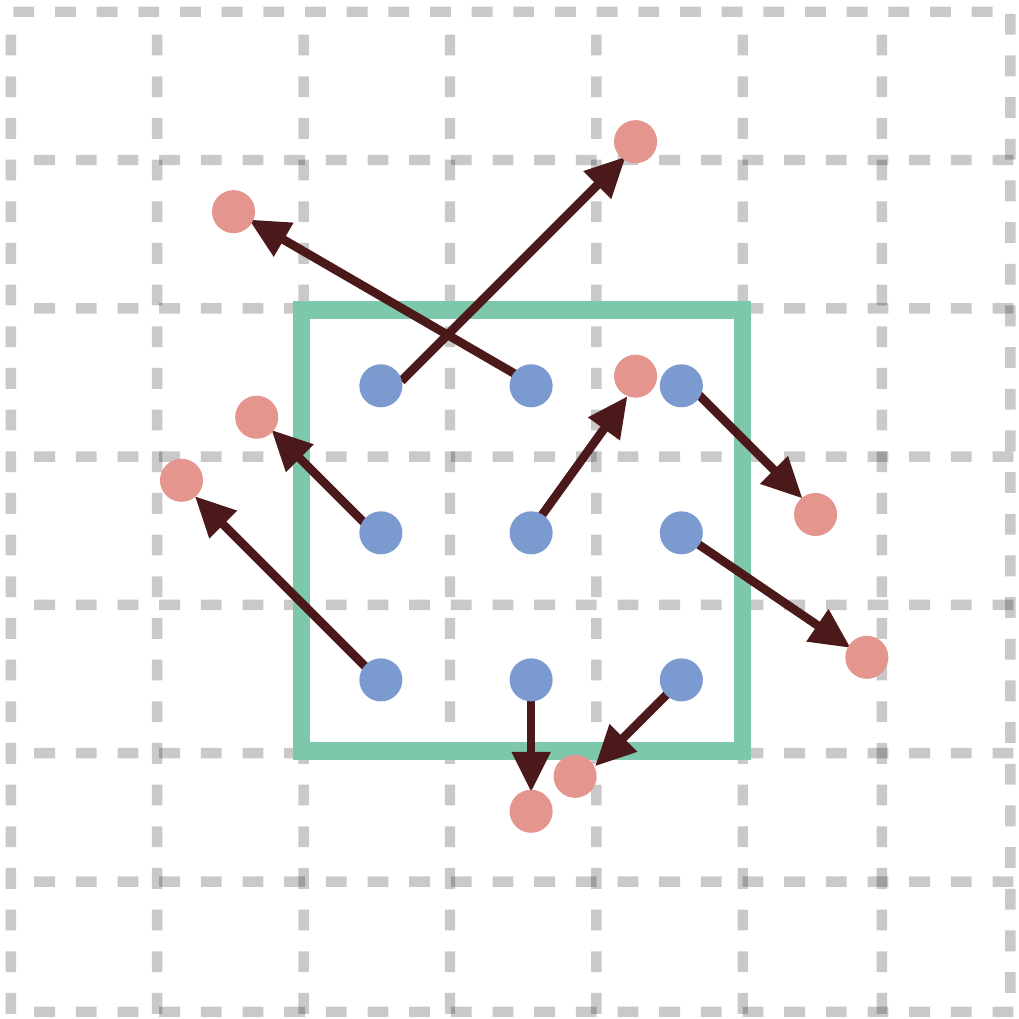}   
    \end{minipage}
  }
  \caption {Illustration of irregular sampling of neighboring pixels using offsets: (a) regular sampling~${\bf{q}}$ on discrete grid; (b) learned offsets~$\Delta {\bf{q}}$; (c) deformable sampling locations~${\bf{s}}({\bf{q}})$ with the offsets~$\Delta {\bf{q}}$. The learned offsets are fractional and the corresponding pixel values are obtained by bilinear interpolation.}
  \label{fig:grid}
\end{figure}

\noindent
\textbf{Weighted average.}
Given the learned kernel~$K$ and sampling offsets~$\Delta {{\bf{q}}}$, we compute the residuals~$\hat f_{\bf{p}} - f_{\bf{p}}$ as a weighted average: 
\vspace{-0.1cm}
\begin{equation}\label{eq:dkn}
\hat f_{\bf{p}}  =  f_{\bf{p}} + \sum_{{\bf{q}} \in \mathcal{N({\bf{p}})}} K_{{\bf{p}}{{\bf{s}}}}(f,g)f_{{\bf{s}}({\bf{q}})},
\vspace{-0.1cm}
\end{equation}
where $\mathcal{N({\bf{p}})}$ is a local $3 \times 3$ window centered at the location $\bf{p}$ on a regular grid~(Fig.~\ref{fig:grid}(a)). We denote by ${{\bf{s}}}({\bf{q}})$ the sampling position computed from the offset~$\Delta {{\bf{q}}}$~(Fig.~\ref{fig:grid}(b)) of the location ${\bf{q}}$ as follows.
\vspace{-0.1cm}
\begin{equation}
{\bf{s}}({\bf{q}}) = {\bf{q}} + \Delta {\bf{q}}.
\vspace{-0.1cm}
\end{equation}
The sampling position ${{\bf{s}}}({\bf{q}})$ predicted by the network is irregular and typically fractional~(Fig.~\ref{fig:grid}(c)). We use a sampler to compute corresponding (sub)~pixel values~$f_{{\bf{s}}({\bf{q}})}$ as 
\vspace{-0.1cm}
\begin{equation}
	f_{{\bf{s}}({\bf{q}})} = \sum_{{\bf{t}} \in \mathcal{R({\bf{s}}({\bf{q}}))}} G({\bf{s}}, {\bf{t}}) f_{{\bf{t}}},
\vspace{-0.1cm}
\end{equation}
where $\mathcal{R({{\bf{s}}}({\bf{q}}))}$ enumerates all integer locations in a local 4-neighborhood system to the fractional position ${\bf{s}}({\bf{q}})$, and $G$ is a sampling kernel. Following~\cite{dai2017deformable,jaderberg2015spatial}, we use a two-dimensional bilinear kernel, and split it into two one-dimensional ones as
\vspace{-0.1cm}
\begin{equation}
	G({\bf{s}}, {\bf{t}}) = g(s_x, t_x) g(s_y, t_y),
\vspace{-0.1cm}
\end{equation}
where $g(a,b) = \operatorname{max}(0,1-|a-b|)$. Note that the residual term in~\eqref{eq:dkn} is exactly the same as the explicit weighted average, but we aggregate pixels from the sparsely chosen locations~${\bf{s}}({\bf{q}})$ with the learned kernels~$K$, which is not feasible in current methods.

When we do not use a residual connection, we compute the upsampling result~$\hat f_{\bf{p}}$ directly as a weighted average using the learned kernels and offsets:
\vspace{-0.1cm}
\begin{equation}\label{eq:dkn_wo_res}
\hat f_{\bf{p}} = \sum_{{\bf{q}} \in \mathcal{N({\bf{p}})}} K_{{\bf{p}}{{\bf{s}}}}(f,g)f_{{\bf{s}}({\bf{q}})}.
\vspace{-0.1cm}
\end{equation}

\noindent
\textbf{Loss.}
We train our model by minimizing the $L_1$ norm of the difference between the network output~$\hat{f}$ and ground-truth HR reference depth map~$f^{\text{gt}}$ as follows.
\vspace{-0.1cm}
\begin{equation}
	L(f^{\text{gt}}, \hat{f}) =  \sum_{\bf{p}}{| f^{\text{gt}}_{\bf{p}} -  \hat f_{\bf{p}} |_{1}}.
\end{equation}
\vspace{-0.3cm}

\noindent
\textbf{Testing.}
Two principles have guided the design of our learning architecture:
(1) Points from a large receptive field in the original guidance and
target images should be used to compute the weighted averages
associated with the value of the upsampled depth map at each one of
its pixels; and (2) inference should be fast. The second principle is
rather self-evident. We believe that the first one is also rather
intuitive, and it is justified empirically by the ablation study
presented later. In fine, it is also the basis for our approach, since
our network learns where, and how to sample a small number of points
in a large receptive field.

A reasonable compromise between receptive field size and speed is to
use one or several convolutional layers with a multi-pixel stride,
which enlarges the image area pixels are drawn from without increasing
the number of weights in the network. This is the approach we have
followed in our base architecture, DKN, with two stride-2 ``DownConv"
layers. The price to pay is a loss in spatial resolution for the final
feature map, with only $1/16$ of the total number $N$ of pixels in the
input images. One could of course give as input to our network the
receptive fields associated with all $N$ of the original guidance and
target image pixels, at the cost of $N$ forward passes during
inference. DKN implements a much more efficient method where $16$
shifted copies of the two images are used in turn as input to the
network, and the corresponding network outputs are then stitched
together in a single HR image, at the cost of only $16$ forward
passes. The details of this {\em shift-and-stitch} approach~\cite{long2015fully,niklaus2017video}
can be found in the supplemental material.

\begin{figure}
\centering
\includegraphics[width=0.8\linewidth]{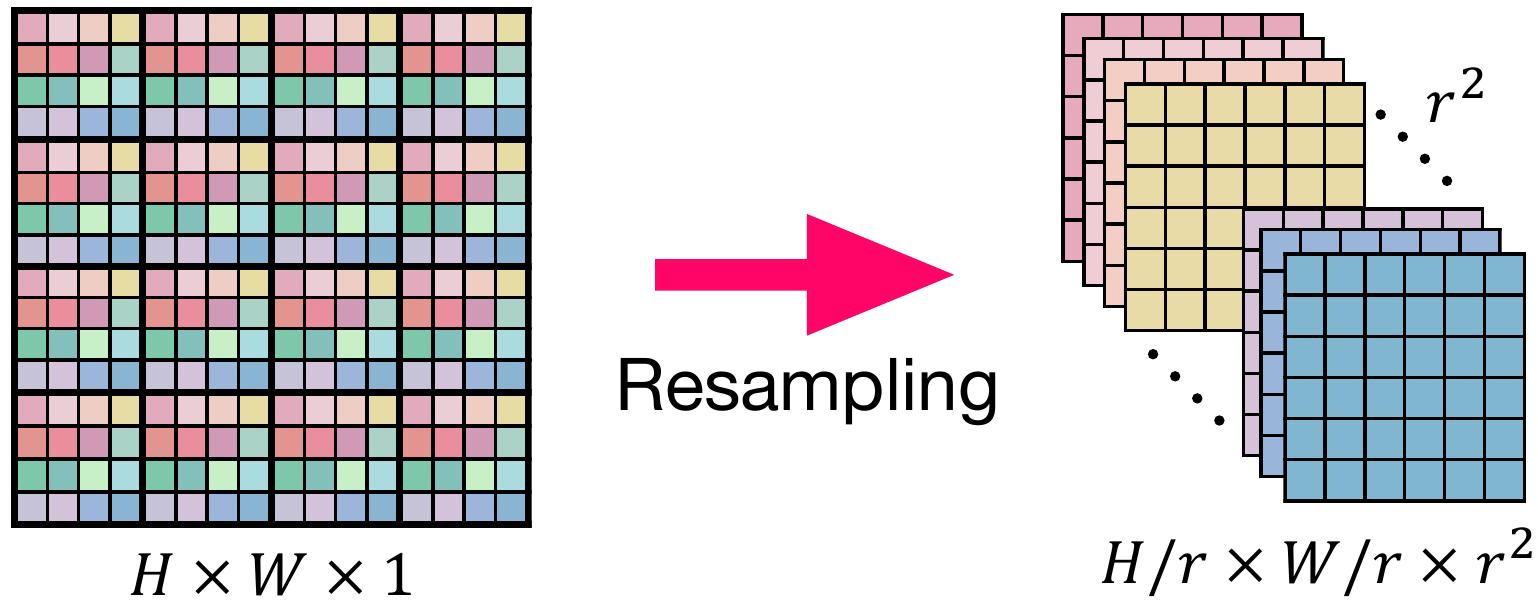}
\caption{Illustration of resampling. An image of size~$H \times W \times 1$ is reshaped with stride~$r$ in each dimension, resulting a resampled one of size~$H/r \times W/r \times r^2$.~(Best viewed in color.)}
\label{fig:resampling}
\vspace{-0.4cm}
\end{figure}

\begin{table*}[t]
\footnotesize
\centering
\caption{Quantitive comparison with the state of the art on depth map upsampling in terms of average RMSE. Numbers in bold indicate the best performance and underscored ones are the second best. Following~\cite{li2016deep,li2017joint}, the average RMSE are measured in centimeter for the NYU v2 dataset~\cite{silberman2012indoor}. For other datasets, we compute RMSE with upsampled depth maps scaled to the range $[0, 255]$.}
\label{table:depth-upsampling}
\addtolength{\tabcolsep}{-4pt}
\renewcommand{\arraystretch}{0.9}
\newcolumntype{L}[1]{>{\raggedright\arraybackslash}p{#1}}
\newcolumntype{C}[1]{>{\centering\arraybackslash}p{#1}}
\newcolumntype{R}[1]{>{\raggedleft\arraybackslash}p{#1}}
\begin{tabular}[c]{L{2.1cm} R{0.75cm} R{0.75cm} R{0.75cm} R{0.75cm} R{0.75cm} R{0.75cm} R{0.75cm} R{0.75cm} R{0.75cm} R{0.75cm} R{0.75cm} R{0.75cm}} 
\midrule
\multicolumn{1}{c}{Datasets} & \multicolumn{3}{c}{Middlebury~\cite{hirschmuller2007evaluation}} & \multicolumn{3}{c}{Lu~\cite{lu2014depth}} & \multicolumn{3}{c}{NYU v2~\cite{silberman2012indoor}} & \multicolumn{3}{c}{Sintel~\cite{butler2012sintel}} \\
\cmidrule(lr){1-13}
\multicolumn{1}{c}{Methods} &  \multicolumn{1}{c}{$4\times$} &  \multicolumn{1}{c}{$8\times$} &  \multicolumn{1}{c}{$16\times$} &  
\multicolumn{1}{c}{$4\times$} &  \multicolumn{1}{c}{$8\times$} &  \multicolumn{1}{c}{$16\times$} & 
\multicolumn{1}{c}{$4\times$} &  \multicolumn{1}{c}{$8\times$} &  \multicolumn{1}{c}{$16\times$} & 
\multicolumn{1}{c}{$4\times$} &  \multicolumn{1}{c}{$8\times$} &  \multicolumn{1}{c}{$16\times$} \\

\cmidrule{1-13}

Bicubic Int. & 4.44 & 7.58 & 11.87 & 5.07 & 9.22 & 14.27 & 8.16 & 14.22 & 22.32 & 6.54 & 8.80 & 12.17 \\
MRF~\cite{diebel2006application} & 4.26 & 7.43 & 11.80 & 4.90 & 9.03 & 14.19 & 7.84 & 13.98 & 22.20 & 8.81 & 11.77 & 15.75 \\
GF~\cite{he2013guided} & 4.01 & 7.22 & 11.70 & 4.87 & 8.85 & 14.09 & 7.32 & 13.62 & 22.03 & 6.10 & 8.22 & 11.22 \\

JBU~\cite{kopf2007joint} & 2.44 & 3.81 & 6.13 & 2.99 & 5.06 & 7.51 & 4.07 & 8.29 & 13.35  & 5.88 & 7.63 & 10.97 \\

TGV~\cite{ferstl2013} & 3.39 & 5.41 & 12.03 & 4.48 & 7.58 & 17.46 & 6.98 & 11.23 & 28.13 & 32.01 & 36.78 & 43.89 \\

Park~\cite{park2011} & 2.82 & 4.08 & 7.26 & 4.09 & 6.19 & 10.14 & 5.21 & 9.56 & 18.10 & 9.28 & 12.22 & 16.51 \\

SDF~\cite{ham2018robust} & 3.14 & 5.03 & 8.83 & 4.65 & 7.53 & 11.52 & 5.27 & 12.31 & 19.24 & 6.52 & 7.98 & 11.36 \\

FBS~\cite{barron2016fast} & 2.58 & 4.19 & 7.30 & 3.03 & 5.77 & 8.48 & 4.29 & 8.94 & 14.59 & 11.96 & 12.29 & 13.08 \\

FGI~\cite{li2016fast}& 3.24 & 4.60 & 6.74 & 4.68 & 6.32 & 9.25 & 6.43 & 9.52 & 14.13 & 6.29 & 8.24 & 11.01 \\
\cmidrule{1-13}
DMSG~\cite{hui2016depth} & 1.88 & 3.45 & 6.28 & 2.30 & 4.17 & 7.22 & 3.02 & 5.38 & 9.17 & 5.32 & 7.24 & 10.11 \\
DJF~\cite{li2016deep}& 2.14 & 3.77 & 6.12 & 2.54 & 4.71 & 7.66 & 3.54 & 6.20 & 10.21 & 5.51 & 7.52 & 10.63 \\
DJFR~\cite{li2017joint}& 1.98 & 3.61 & 6.07 & 2.22 & 4.54 & 7.48 & 3.38 & 5.86 & 10.11 & 5.50 & 7.43 & 10.48 \\
\cmidrule{1-13}\morecmidrules\cmidrule{1-13}

FDKN$^\dagger$ & \textbf{1.07} & 2.23 & 5.09 & \underline{0.85} & \underline{1.90} & 5.33 & 2.05 & 4.10 & 8.10 & \underline{3.31} & 5.08 & 8.51\\

DKN$^\dagger$ & 1.12 & \underline{2.13} & 5.00 & 0.90 & \textbf{1.83} & \textbf{4.99} & 2.11 & 4.00 & 8.24 & 3.40 & 4.90 & 8.18 \\
\cmidrule{1-13}
FDKN w/o Res. & 1.12 & 2.23 & 4.52 & \underline{0.85} & 2.19 & 5.15 & 1.88 & 3.67 & 7.13 & 3.38 & 5.02 & 7.74 \\

DKN w/o Res. & 1.26 & 2.16 & \underline{4.32} & 0.99 & 2.21 & 5.12 & \underline{1.66} & \underline{3.36} & \underline{6.78} & 3.36 & \underline{4.82} & \textbf{7.48} \\

FDKN & \underline{1.08} & 2.17 & 4.50 & \textbf{0.82} & 2.10 & \underline{5.05} & 1.86 & 3.58 & 6.96 & 3.36 & 4.96 & 7.74 \\

DKN& 1.23 & \textbf{2.12} & \textbf{4.24} & 0.96 & 2.16 & 5.11 & \textbf{1.62} & \textbf{3.26} & \textbf{6.51} & \textbf{3.30} & \textbf{4.77} & \underline{7.59} \\
\midrule
\end{tabular}
\vspace{-0.4cm}
\end{table*}

\vspace{-0.2cm}
\subsection{FDKN architecture}
\vspace{-0.2cm}
A more efficient alternative to DKN is to split the input images into
the same 16 subsampled and shifted parts as before, but this time {\em
  stack} them into new target and guidance images~(Fig.~\ref{fig:resampling}), with $16$ channels
for the former, and $16C$ channels for the latter, \eg,~$C=3$ when the RGB image is
used. The effective receptive field for FDKN is comparable to
that of DKN, but FDKN involves much fewer parameters because of the
reduced input image resolution and the shared weights across
channels. The individual channels are then recomposed into the final
upsampled image~\cite{shi2016real}, at the cost of only one forward pass. Specifically, we use a series of 6 convolutional layers of size~$3 \times 3$ for feature extraction. For weight and offset regression, we apply a $1\times 1$ convolution on top of the feature extraction layers similar to DKN, but using more network parameters. For example, FDKN and DKN compute feature maps of size $16k^2 \times 1 \times 1$ and $k^2 \times 1 \times 1$, respectively, for weight regression, from each feature of size $128 \times 1 \times 1$. This allows FDKN to estimate kernel weights and offsets for all pixels simultaneously. The details of this {\em shift-and-stack} approach can be found in the supplemental material. In practice, FDKN gives a 17 times speed-up over DKN. Because it involves fewer parameters~($0.6$M vs. $1.1$M for DKN), one might expect somewhat degraded results. Our experiments demonstrate that FDKN remains in the ballpark of that of DKN, still significantly better than competing approaches, and in one case even overperforming DKN.

\vspace{-0.2cm}
\section{Experiments}\label{sec:exp}
\vspace{-0.2cm}
In this section we present a detailed analysis and evaluation of our approach. More results and other applications of our model including saliency image upsampling, cross-modality image restoration, texture removal and semantic segmentation can be found in the supplement. 
\vspace{-0.2cm}
\subsection{Implementation details}
\vspace{-0.2cm}

Following the experimental protocol of~\cite{li2016deep,li2017joint}, we train different models to upsample depth maps for scale factors $4 \times$, $8 \times$, $16 \times$ with random initialization. We sample 1,000 RGB/D image pairs of size $640 \times 480$ from the NYU v2 dataset~\cite{silberman2012indoor}. We use the same image pairs as in~\cite{li2016deep,li2017joint} to train the networks. The models are trained with a batch size of 1 for 40k iterations, giving roughly 20 epochs over the training data. We synthesize LR depth images~($4 \times$, $8 \times$, $16 \times$)~from ground truth by bicubic downsampling. We use the Adam optimizer~\cite{kingma2015adam} with $\beta_1 = 0.9$ and $\beta_2 = 0.999$. As learning rate we use $0.001$ and divide it by 5 every 10k iterations. Data augmentation and regularization techniques such as weight decay and dropout~\cite{krizhevsky2012imagenet} are not used, since 1,000 RGB/D image pairs from the NYU dataset have proven to be sufficient to train our models (See the supplement). All networks are trained end-to-end using \texttt{PyTorch}~\cite{paszke2017automatic}.

\begin{figure*}[t]  
  \centering
  \footnotesize	
\captionsetup[subfloat]{labelformat=empty}
  \subfloat[{RGB image.}]{
    \begin{minipage}[b]{0.146\linewidth} 
      \includegraphics[width=\linewidth, height=1.8975cm]{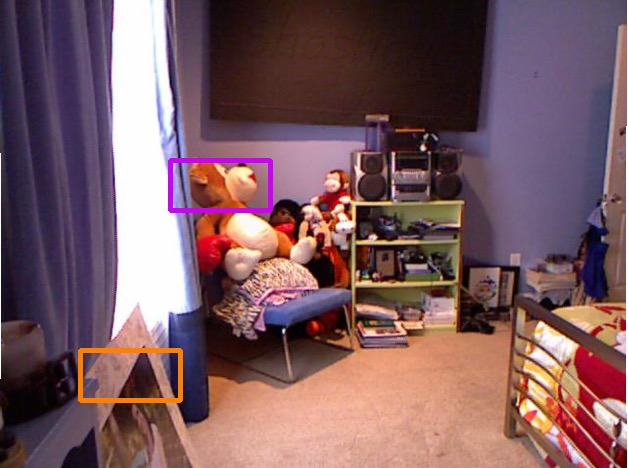}\vspace{0.05cm} 
      \includegraphics[width=\linewidth, height=1.8975cm]{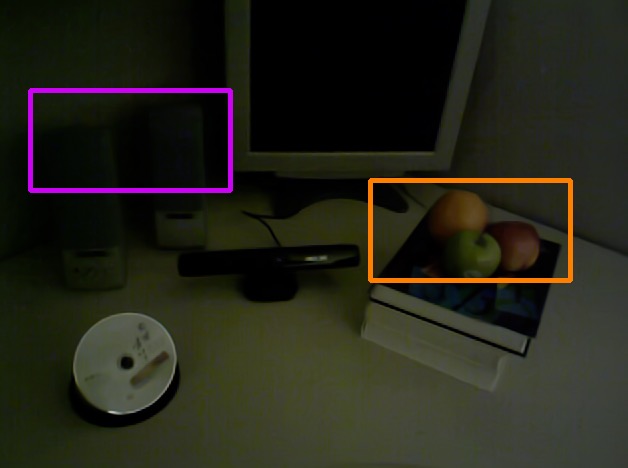}\vspace{0.05cm}
      \includegraphics[width=\linewidth, height=1.8975cm]{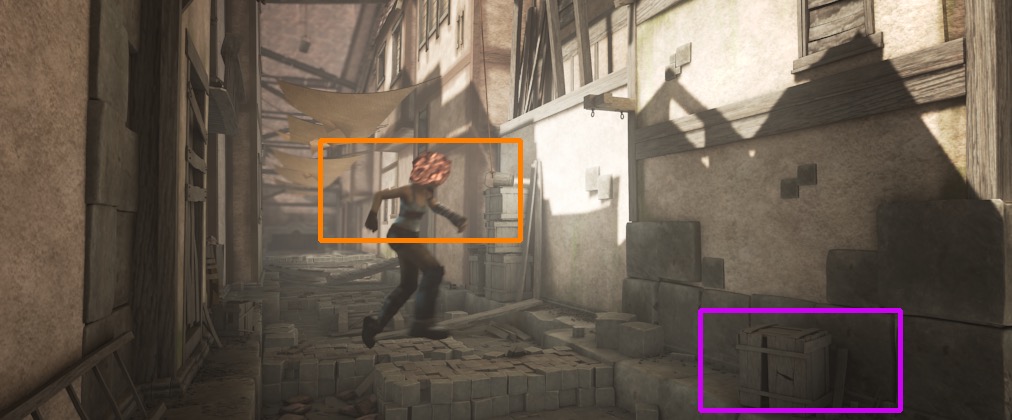}
    \end{minipage}
  }\hspace{-0.2cm}
  \subfloat[{GF~\cite{he2013guided}}.]{
    \begin{minipage}[b]{0.106\linewidth} 
      \includegraphics[width=\linewidth]{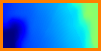}
      \includegraphics[width=\linewidth]{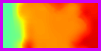}\vspace{0.05cm}    
      \includegraphics[width=\linewidth]{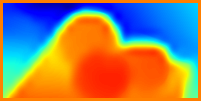}
      \includegraphics[width=\linewidth]{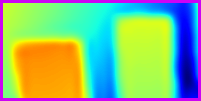}\vspace{0.05cm}
      \includegraphics[width=\linewidth]{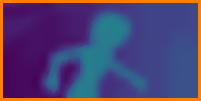}
      \includegraphics[width=\linewidth]{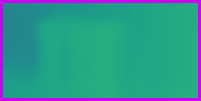}
    \end{minipage}
  }\hspace{-0.2cm}
  \subfloat[{TGV~\cite{ferstl2013}}.]{
    \begin{minipage}[b]{0.106\linewidth} 
      \includegraphics[width=\linewidth]{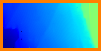}
      \includegraphics[width=\linewidth]{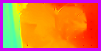}\vspace{0.05cm}
      \includegraphics[width=\linewidth]{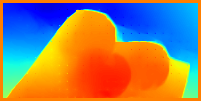}
      \includegraphics[width=\linewidth]{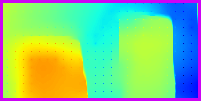}\vspace{0.05cm} 
      \includegraphics[width=\linewidth]{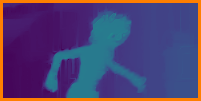}
      \includegraphics[width=\linewidth]{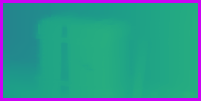}
    \end{minipage}
  }\hspace{-0.2cm}
  \subfloat[{Park~\cite{park2011}}.]{
    \begin{minipage}[b]{0.106\linewidth} 
      \includegraphics[width=\linewidth]{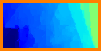}
      \includegraphics[width=\linewidth]{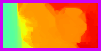}\vspace{0.05cm}    
      \includegraphics[width=\linewidth]{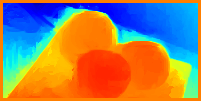}
      \includegraphics[width=\linewidth]{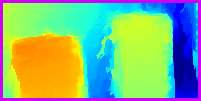}\vspace{0.05cm} 
      \includegraphics[width=\linewidth]{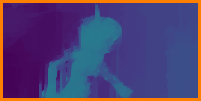}
      \includegraphics[width=\linewidth]{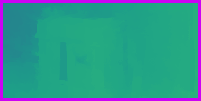}
    \end{minipage}
  }\hspace{-0.2cm}
  \subfloat[{SDF~\cite{ham2018robust}}.]{
    \begin{minipage}[b]{0.106\linewidth} 
      \includegraphics[width=\linewidth]{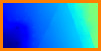}
      \includegraphics[width=\linewidth]{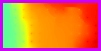}\vspace{0.05cm}
      \includegraphics[width=\linewidth]{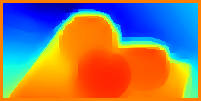}
      \includegraphics[width=\linewidth]{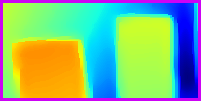}\vspace{0.05cm}
      \includegraphics[width=\linewidth]{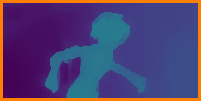}
      \includegraphics[width=\linewidth]{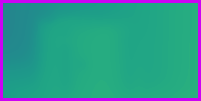}
    \end{minipage}
  }\hspace{-0.2cm}
  \subfloat[{DJFR~\cite{li2017joint}}.]{
    \begin{minipage}[b]{0.106\linewidth} 
      \includegraphics[width=\linewidth]{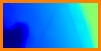}
      \includegraphics[width=\linewidth]{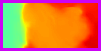}\vspace{0.05cm}
      \includegraphics[width=\linewidth]{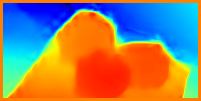}
      \includegraphics[width=\linewidth]{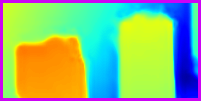}\vspace{0.05cm} 
      \includegraphics[width=\linewidth]{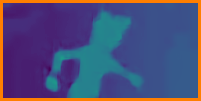}
      \includegraphics[width=\linewidth]{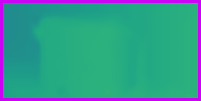}
    \end{minipage}
  }\hspace{-0.2cm}
  \subfloat[{DKN}.]{
    \begin{minipage}[b]{0.106\linewidth} 
      \includegraphics[width=\linewidth]{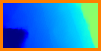}
      \includegraphics[width=\linewidth]{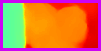}\vspace{0.05cm}
      \includegraphics[width=\linewidth]{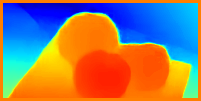}
      \includegraphics[width=\linewidth]{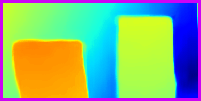}\vspace{0.05cm} 
      \includegraphics[width=\linewidth]{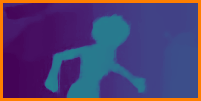}
      \includegraphics[width=\linewidth]{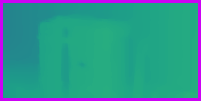}
    \end{minipage}
  }\hspace{-0.2cm}
  \subfloat[{FDKN}.]{
    \begin{minipage}[b]{0.106\linewidth} 
      \includegraphics[width=\linewidth]{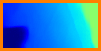}
      \includegraphics[width=\linewidth]{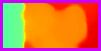}\vspace{0.05cm}
      \includegraphics[width=\linewidth]{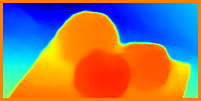}
      \includegraphics[width=\linewidth]{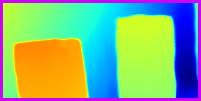}\vspace{0.05cm} 
      \includegraphics[width=\linewidth]{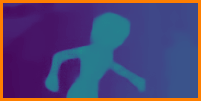}
      \includegraphics[width=\linewidth]{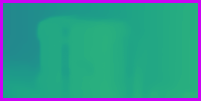}
    \end{minipage}
  }\hspace{-0.2cm}
  \subfloat[{Ground truth}.]{
    \begin{minipage}[b]{0.106\linewidth} 
      \includegraphics[width=\linewidth]{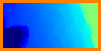}
      \includegraphics[width=\linewidth]{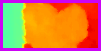}\vspace{0.05cm}
      \includegraphics[width=\linewidth]{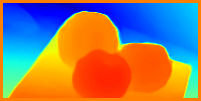}
      \includegraphics[width=\linewidth]{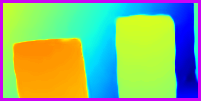}\vspace{0.05cm} 
      \includegraphics[width=\linewidth]{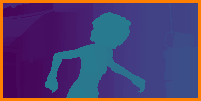}
      \includegraphics[width=\linewidth]{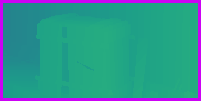}
    \end{minipage}
  }\hspace{-0.2cm}
  \caption{Visual comparison of upsampled depth maps~($8 \times$). Top to bottom: Each row shows upsampled depth maps on the NYU v2~\cite{silberman2012indoor}, Lu~\cite{lu2014depth}, and Sintel~\cite{butler2012sintel} datasets, respectively. Note that we train our models with the NYU v2 dataset, and do not fine-tune them to other datasets.~(Best viewed in color.)}
\label{fig:depth_upsampling}
\vspace{-0.3cm}
\end{figure*}

\vspace{-0.2cm}
\subsection{Results}
\vspace{-0.2cm}
We test our models with the following four benchmark datasets. These feature aligned color and depth images. Note that we train our models with the NYU v2 dataset, and do not fine-tune them to other ones to evaluate its generalization ability.

\begin{itemize}[leftmargin=*]
	\vspace{-0.2cm}
	\item[$\bullet$] Middlebury dataset~\cite{hirschmuller2007evaluation,scharstein2007learning}: We use the 30 RGB/D image pairs from the 2001-2006 datasets provided by Lu~\cite{lu2014depth}.
		\vspace{-0.2cm}
	\item[$\bullet$] Lu dataset~\cite{lu2014depth}: This provides 6 RGB/D image pairs acquired by the ASUS Xtion Pro camera~\cite{asus2018xtion}. 
		\vspace{-0.2cm}
	\item[$\bullet$] NYU v2 dataset~\cite{silberman2012indoor}: It consists of 1,449 RGB/D image pairs captured with the Microsoft Kinect~\cite{zhang2012kinect}. We exclude the 1,000 pairs used for training, and use the rest (449 pairs) for evaluation
		\vspace{-0.2cm}
	\item[$\bullet$] Sintel dataset~\cite{butler2012sintel}: This dataset provides 1,064 RGB/D image pairs created from an animated 3D movie. It contains realistic scenes including fog and motion blur. We use 864 pairs from a final-pass dataset for testing.
		\vspace{-0.2cm}
\end{itemize}

\begin{table*}[http]
\footnotesize
\centering
\caption{Runtime comparison for HR images of size~$640 \times 480$~(NYU v2 dataset~\cite{silberman2012indoor}).}
\label{table:runtime}
\addtolength{\tabcolsep}{-4.0pt}
\newcolumntype{L}[1]{>{\raggedright\arraybackslash}p{#1}}
\newcolumntype{C}[1]{>{\centering\arraybackslash}p{#1}}
\newcolumntype{R}[1]{>{\raggedleft\arraybackslash}p{#1}}
\begin{tabular}[c]{L{1.7cm} C{0.8cm} C{0.8cm} C{0.8cm} C{0.8cm} C{0.8cm} C{0.8cm} C{0.8cm} C{0.8cm} C{0.8cm} C{0.8cm} C{0.8cm} C{0.8cm} C{0.8cm} C{0.8cm} C{0.8cm} C{0.8cm}} 
\hline
\multicolumn{1}{c}{} &  \multicolumn{1}{c}{MRF~\cite{diebel2006application}} & \multicolumn{1}{c}{GF~\cite{he2013guided}} &  \multicolumn{1}{c}{JBU~\cite{kopf2007joint}} &  
\multicolumn{1}{c}{TGV~\cite{ferstl2013}} &  \multicolumn{1}{c}{Park~\cite{park2011}} &  \multicolumn{1}{c}{SDF~\cite{ham2018robust}} & 
\multicolumn{1}{c}{FBS~\cite{barron2016fast}} &  \multicolumn{1}{c}{FGI~\cite{li2016fast}} & \multicolumn{1}{c}{DMSG~\cite{hui2016depth}} &  \multicolumn{1}{c}{DJFR~\cite{li2017joint}} & \multicolumn{1}{c}{DKN$^\dagger$} & \multicolumn{1}{c}{FDKN$^\dagger$} & \multicolumn{1}{c}{DKN} & \multicolumn{1}{c}{FDKN}\\
\hline
\multicolumn{1}{c}{Times~(s)} & 0.69 & 0.14 & 0.31 & 33 & 18 & 25 & 0.37 & 0.24 & \underline{0.04} & {\bf 0.01} & 0.09 & \bf{0.01} & 0.17 & \bf{0.01} \\
\hline
\end{tabular}
\vspace{-0.2cm}
\end{table*}

\begin{table*}[http]
\footnotesize
\centering
\caption{Average RMSE comparison~(DKN/FDKN) of different components and size of kernels~(from $3 \times 3$ to $25 \times 25$). From the third row, we can see that aggregating pixels from a $15 \times 15$ window is enough. We thus restrict the maximum range of offset locations to $15 \times 15$. For example, results for $7 \times 7$ in the forth row are computed using $49$ pixels sparsely sampled from a $15 \times 15$ window. We omit the results for $15\times 15$, $19\times 19$ and $25\times 25$ kernels, since they are equal to or beyond the maximum range of offset locations.}
\label{table:ablation-archicture}
\addtolength{\tabcolsep}{-3.7pt}
\renewcommand{\arraystretch}{0.5}
\newcolumntype{C}[1]{>{\centering\arraybackslash}p{#1}}
\begin{tabular}[c]{c c c c c C{0.6cm}@{/} C{0.6cm} C{0.6cm}@{/} C{0.6cm} C{0.6cm}@{/} C{0.6cm} C{0.6cm}@{/} C{0.6cm} C{0.6cm}@{/} C{0.6cm} C{0.6cm}@{/} C{0.6cm}} 
\midrule
\multicolumn{2}{c}{Weight learning}  & \multicolumn{2}{c}{Offset learning} & \multirow{2}{*}{Res.} & \multicolumn{2}{c}{\multirow{2}{*}{$3 \times 3$}} & \multicolumn{2}{c}{\multirow{2}{*}{$5 \times 5$}} & \multicolumn{2}{c}{\multirow{2}{*}{$7 \times 7$}} & \multicolumn{2}{c}{\multirow{2}{*}{$15 \times 15$}} & \multicolumn{2}{c}{\multirow{2}{*}{$19 \times 19$}} & \multicolumn{2}{c}{\multirow{2}{*}{$ 25 \times 25$}}\\
\cmidrule(lr){1-2} \cmidrule(lr){3-4}  RGB  & Depth  &  RGB  & Depth   \\ 
\cmidrule{1-17}
$\checkmark$ & & & & & 5.92 & 6.05 & 5.52 & 5.73 & 5.43 & 5.67 & 5.59 & 5.74 & 5.82 & 5.81 & 6.21 &5.99 \\
\cmidrule{1-17}
& $\checkmark$ & & & & 5.24 & 5.30 & 4.36 & 4.47 & 4.09 & 4.24 & 4.09 & 4.17 & 4.11 & 4.18 & 4.15 &4.21 \\
\cmidrule{1-17}
$\checkmark$ & $\checkmark$ &  & & & 5.03 & 5.14 & 3.90 & 4.16 & 3.48 & 3.80 & 3.32 & 3.66 & 3.33 & 3.66 & 3.39 & 3.72 \\
\cmidrule{1-17}
$\checkmark$ & & $\checkmark$ & & & 5.37 & 5.18 & 5.38 & 5.09 & 5.40 & 5.07 & \multicolumn{2}{c}{--} & \multicolumn{2}{c}{--} & \multicolumn{2}{c}{--} \\
\cmidrule{1-17}
& $\checkmark$ & &$\checkmark$ & & 4.06 & 4.13 & 4.09 & 4.13 & 4.13 & 4.14 & \multicolumn{2}{c}{--} & \multicolumn{2}{c}{--} & \multicolumn{2}{c}{--} \\
\cmidrule{1-17}

\cmidrule{1-17}
$\checkmark$ &$\checkmark$ & $\checkmark$ & $\checkmark$& & 3.36 & 3.67 & 3.32 & 3.65 & 3.33 & 3.66 & \multicolumn{2}{c}{--} & \multicolumn{2}{c}{--} & \multicolumn{2}{c}{--} \\
\cmidrule{1-17}
$\checkmark$ &$\checkmark$ & $\checkmark$ & $\checkmark$&$\checkmark$ & 3.26 & 3.58 & \underline{3.21} & \underline{3.53} & \textbf{3.19} & \textbf{3.52} & \multicolumn{2}{c}{--} & \multicolumn{2}{c}{--} & \multicolumn{2}{c}{--} \\
\midrule
\end{tabular}
\vspace{-0.4cm}
\end{table*}

We compare our method with the state of the art in Table~\ref{table:depth-upsampling}.~It shows the average RMSE between upsampling results and ground truth. All numbers except those for the Sintel dataset are taken from~\cite{li2016deep,li2017joint}. The results of DJF~\cite{li2016deep} and its residual version~(DJFR~\cite{li2017joint}) are obtained by the provided models trained with the NYU v2 dataset. DMSG~\cite{hui2016depth} uses the Middlebury and Sintel datasets for training the network. For fair comparison of DMSG with other CNN-based methods including ours, we retrain the DMSG model using the same image pairs from the NYU v2 dataset as in~\cite{li2016deep,li2017joint}. From this table, we observe four things:~(1) Our models outperform the state of the art including CNN-based methods~\cite{hui2016depth,li2016deep,li2017joint} by significant margins in terms of RMSE, even without the residual connection~(DKN w/o Res. and FDKN w/o Res.). For example, DKN decreases the average RMSE by $52\%$~($4 \times$), $44\%$~($8 \times$) and $36\%$~($16 \times$) compared to DJFR~\cite{li2017joint}. (2)~Our models trained \emph{without the guidance of HR color images}~(DKN$^\dagger$ and FDKN$^\dagger$), using the depth map only, also outperform the state of the art. In particular, they give better results than DKN and FDKN for the Lu dataset~\cite{lu2014depth}. A plausible expiation is that depth and color boundaries are less correlated, since the color images in the dataset are captured in a low-light condition. (3)~We can clearly see that our models perform well on both synthetic and real datasets~(\eg, the Sintel and NYU v2 datasets), and generalize well to other images~(\eg, on the Middlebury dataset)~outside the training dataset. (4)~FDKN retains the superior performance of DKN, and even outperforms DKN for the Lu dataset.

\noindent{\textbf{Qualitative results.}}
Figure~\ref{fig:depth_upsampling} shows a visual comparison of the upsampled depth maps~(8$\times$). The better ability to extract common structures from the color and depth images by our models here is clearly visible. Specifically, our results show a sharp depth transition without the texture-copying artifacts. In contrast, artifacts are clearly visible even in the results of DJFR~\cite{li2017joint}, which tends to over-smooth the results and does not recover fine details. This confirms once more the advantage of using the weighted average with spatially-variant kernels and an adaptive neighborhood system in depth map upsampling.

\noindent{\textbf{Runtime.}} Table~\ref{table:runtime} shows runtime comparisons on the same machine. We report the runtime for DMSG~\cite{hui2016depth}, DJFR~\cite{li2017joint} and our models with a Nvidia Titan XP and for other methods with an Intel i5 3.3 GHz CPU. Our current implementation for DKN takes on average $0.17$ seconds for HR images of size~$640 \times 480$. It is slower than DMSG~\cite{hui2016depth} and DJFR~\cite{li2017joint}, but yields a significantly better RMSE~(Fig.~\ref{fig:timegraph} and Table~\ref{table:depth-upsampling}). FDKN runs about $17 \times$ faster than the DKN, as fast as DJFR, but with significantly higher accuracy.

\vspace{-0.2cm}
\subsection{Discussion}\label{sec:discussion}
\vspace{-0.2cm}
We conduct an ablation analysis on different components in our models, and show the effects of different parameters for depth map upsampling~($8 \times$) on the NYU v2 dataset~\cite{silberman2012indoor}. More discussion can be found in the supplement.

\noindent{\textbf{Network architecture.}}
We show the average RMSE for six variants of our models in Table~\ref{table:ablation-archicture}. The baseline models learn kernel weights from HR color images only. The first row shows that this baseline already outperforms the state of the art~(see Table~\ref{table:depth-upsampling}). From the second row, we can see that our models trained using LR depth maps only give better results than the baseline, indicating that using the HR color images only is not enough to fully exploit common structures. The third row demonstrates that constructing kernels from both images boosts performance. For example, the average RMSE of DKN decreases from $5.92$ to $5.03$ for the $3\times 3$ kernel. The fourth and fifth rows show that learning the offsets significantly boosts the performance of our models. The average RMSE of DKN trained using the HR color or LR depth images only decreases from $5.92$ to $5.37$ and from $5.24$ to $4.06$, respectively, for the $3\times 3$ kernel. The last two rows demonstrate that the effect of learning kernel weights and offsets from both inputs is significant, and combining all components including the residual connection gives the best results. Note that learning to predict the spatial offsets is important because (1)~learning spatially-variant kernels for individual pixels would be very hard otherwise, unless using much larger kernels to achieve the same neighborhood size, which would lead to an inefficient implementation, and (2) contrary to current architectures including DJF~\cite{li2016deep} and DMSG~\cite{hui2016depth}, this allows sub-pixel information aggregation.

\noindent{\textbf{Kernel size.}}
Table~\ref{table:ablation-archicture} also compares the performances of networks with different size of kernels. We enlarge the kernel size gradually from $3 \times 3$ to $25 \times 25$ and compute the average RMSE. From the third row, we observe that the performance improves until size of $15 \times 15$. Increasing size further does not give additional performance gain. This indicates that aggregating pixels from a $15 \times 15$ window is enough for the task. For offset learning, we restrict the maximum range of the sampling position to $15\times 15$ for all experiments. That is, the results from the third to last rows are computed by aggregating 9, 25 or 49 samples sparsely chosen from a $15 \times 15$ window. The last row of Table~\ref{table:ablation-archicture} suggests that our final models also benefit from using more samples. The RMSE for DKN decreases from $3.26$ to $3.19$ at the cost of additional runtime. For comparison, DKN with kernels of size $3 \times 3$, $5 \times 5$ and $7 \times 7$ take 0.17, 0.18 and 0.19 seconds, respectively, with a Nvidia Titan XP. A $3 \times 3$ size offers a good compromise in terms of RMSE and runtime and this is what we have used in all experiments.  

\noindent{\textbf{DownConv for DKN.}}
We empirically find that extracting features from large receptive fields is important to incorporate context for weight and offset learning. For example, reducing the size from $51 \times 51$ to $23 \times 23$ causes an increase of the average RMSE from $3.26$ to $5.00$ for the $3\times 3$ kernel. The DKN without DownConv layers can be implemented in a single forward pass, but requires more parameters~($1.6$M vs. $1.1$M for DKN) to maintain the same receptive field size, with a total number of convolutions increasing from $0.6$M to $1$M at each pixel. We may use dilated convolutions~\cite{yu2016multi} that support large receptive fields without loss of resolution. When using the same receptive field size as $51 \times 51$, the average RMSE for dilated convolutions increases from $3.26$ to $4.30$ for the $3\times 3$ kernel. The resampling technique~(Fig.~\ref{fig:resampling}) thus appears to be the preferable alternative.

\vspace{-0.2cm}
\section{Conclusion}
\vspace{-0.2cm}
We have presented a CNN architecture for depth map upsampling. Instead of regressing the upsampling results directly from the network, we use spatially-variant weighted averages where the set of neighbors and the corresponding kernel weights are learned end-to-end. A fast version achieves a $17 \times$ speed-up compared to the plain DKN without much~(if any) loss in performance. Finally, we have shown that the weighted averaging process, even using the LR depth image only without any guidance, with $9$ samples sparsely chosen, is sufficient to set a new state of the art.

\clearpage

{\small
\bibliographystyle{ieee}
\bibliography{egbib}
}




\section*{Supplementary material}

# Deformable kernel networks for guided depth map upsampling
# Supplement

Beomjun Kim
Yonsei University

Jean Ponce
Inria / PSL Research University

Bumsub Ham
Yonsei University

Here we present a detailed description of efficient implementation of DKN and the network structure of FDKN in Sec. 1. We then discuss other issues including kernel visualization, the number of feature channels, DownConv for DKN, upscaling factors for training/testing, and kernel prediction vs. direct regression in Sec. 2. We show more results and other applications of our model including saliency image upsampling, cross-modality image restoration, texture removal, and semantic segmentation in Secs. 3 and 4, respectively.

## 1. Network architecture

### 1.1. Efficient implementation for DKN

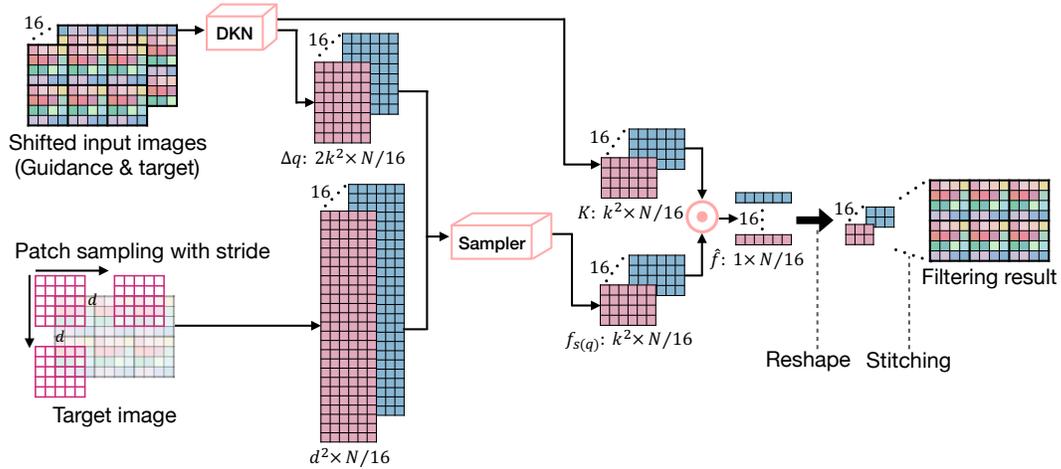

Figure 1. Efficient implementation using a shift-and-stitch approach. We denote $d$ by the maximum range of the sampling location $\mathbf{s}(\mathbf{q})$. We shift the input images and compute the filtering result for each shifted input. We then stitch them up to get a final output that has the same resolution as the inputs. Our approach makes it possible to reuse the storage for kernel weights, offsets, and resampled pixels. See text for details. (Best viewed in color.)

We use the shift-and-stitch approach [16, 19] that stitches the network outputs from shifted versions of the input (Fig. 1). We can obtain the same result as the pixel-wise implementation in 16 forward passes. We first shift input images $x$ pixels to the left and $y$ pixels up, once for every $(x, y)$ where $\{(x, y) | 0 \leq x, y \leq 3\}$, and obtain a total of 16 shifted inputs. Each shifted input goes through the network that gives the kernel weights $K$ and the offsets $\Delta \mathbf{q}$ of size $k^2 \times N/16$ and $2k^2 \times N/16$, respectively. The next step is to obtain image values $f_{\mathbf{s}(\mathbf{q})}$ using the sampling function $\mathbf{s}(\mathbf{q})$ from the target image. To this end, starting from every location $(x, y)$ in the target image, we sample patches of size $d \times d$ with stride 4 in each dimension, each of which gives the output of size $d^2 \times N/16$. The patch size corresponds to the maximum range of the sampling position $\mathbf{s}(\mathbf{q})$. For an efficient implementation, we restrict the range (e.g., to $15 \times 15$ in our experiment). We then sample $k^2$ pixels using the sampling position $\mathbf{s}(\mathbf{q})$ from patches of size $d \times d$, obtaining $f_{\mathbf{s}(\mathbf{q})}$ of size $k^2 \times N/16$ for each shifted input. To compute a weighted average, we apply element-wise multiplication between the kernel weights $K$ and the corresponding sampled pixels $f_{\mathbf{s}(\mathbf{q})}$ of size $k^2 \times N/16$ followed by column-wise summation, resulting in an output of size

1

$1 \times N/16$. Finally, we stitch 16 outputs of size $1 \times N/16$ into a single one to get the final output. Note that one can stitch kernel weights and offsets first and then compute a weighted average, but this requires a large amount of memory. We stitch instead the outputs after the weighted average, and reuse the storage for kernel weights, offsets, and sampled pixels.

## 1.2. FDKN architecture

We show in Fig. 2 the architecture of the FKDN and in Table 1 its detailed description, respectively.

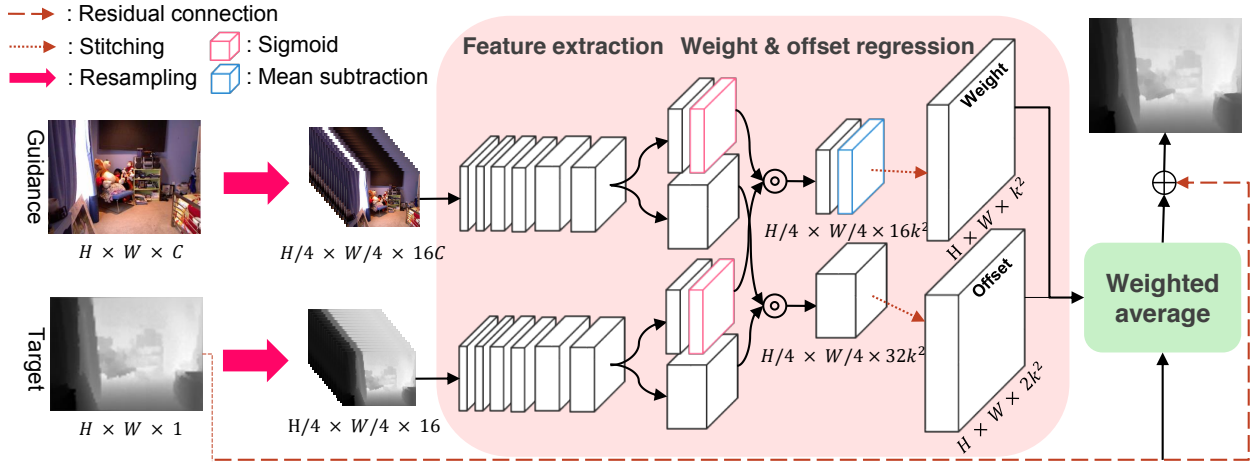

Figure 2. The FDKN architecture. We resample a HR color image of size $H \times W \times C$ with stride 4 in each dimension, where $H$, $W$ and $C$ are height, width and the number of channels, respectively. This gives resampled HR color images of size $H/4 \times W/4 \times 16C$. The depth map is also resampled to the size of $H/4 \times W/4 \times 16$. This allows the FDKN to maintain a receptive field size comparable with the DKN. The FDKN inputs resampled HR color and LR depth images, and then computes the kernel weight and offset locations of size $H \times W \times k^2$ and $H \times W \times 2k^2$, respectively, making it possible to get upsampling results in a single forward pass without loss of resolution. Finally, the residuals computed by the weighted average and the target image are combined to obtain the upsampling result. See Table 1 for the detailed description of the network structure. (Best viewed in color.)

Table 1. FDKN architecture details. The inputs of FDKN are $16C$-channel HR color and 16-channel LR depth images (denoted by $D$). Note that the receptive field of size $13 \times 13$ in the resampled images is comparable to that of size $51 \times 51$ in the DKN.

| Feature extraction | | Weight regression | |
| --- | --- | --- | --- |
| Type | Output | Type | Output |
| Input | $D \times 13 \times 13$ | Conv($1 \times 1$) | $16k^2 \times 1 \times 1$ |
| Conv($3 \times 3$)-BN-ReLU | $32 \times 11 \times 11$ | Sigmoid | $16k^2 \times 1 \times 1$ |
| Conv($3 \times 3$)-ReLU | $32 \times 9 \times 9$ | Mean subtraction or | |
| Conv($3 \times 3$)-BN-ReLU | $64 \times 7 \times 7$ | L1 norm. (w/o Res.) | $16k^2 \times 1 \times 1$ |
| Conv($3 \times 3$)-ReLU | $64 \times 5 \times 5$ | Offset regression | |
| Conv($3 \times 3$)-BN-ReLU | $128 \times 3 \times 3$ | Type | Output |
| Conv($3 \times 3$)-ReLU | $128 \times 1 \times 1$ | Conv($1 \times 1$) | $32k^2 \times 1 \times 1$ |

## 2. Discussion

In this section, we show qualitative results for an ablation analysis on different components in our models, and show the effects of different parameters for depth map upsampling ($\times 8$) on the NYU v2 dataset [21]. We also discuss other issues including kernel visualization and upscaling factors for training and testing.

**Network architecture.** Figure 3 shows a visual comparison of using different networks for depth map upsampling. See Table 4 in the main paper for a quantitive comparison.

**Feature channels.** In Table 2, we compare the effects of the number of feature channels in terms of RMSE, runtime, the number of network parameters, and model size. We use our DKN and FDKN models including the residual connection and a fixed size of $3 \times 3$ kernels. We vary the number of channels $n_i$ ($i = 1, 2$) in the final two layers for feature extraction

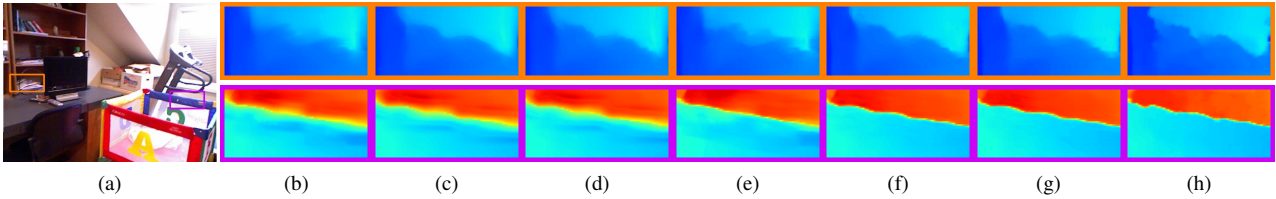

(a) (b) (c) (d) (e) (f) (g) (h)

Figure 3. Visual comparison of different networks (DKN) for depth upsampling (8×) using the kernel of size 3 × 3. (a) RGB image. Results for weight regression using (b) RGB, (c) depth and (d) RGB and depth images. Results for weight and offset regression using (e) RGB, (f) RGB and depth images, and (g) RGB and depth images with the residual connection. (h) Ground truth.

Table 2. Quantitative comparison of using different number of channels for feature extraction. The results of DKN and FDKN are separated by "/". We denote by $n_i$ ($i = 1, 2$) the number of channels in the final two layers for feature extraction. For each network, numbers in bold indicate the best performance and underscored ones are the second best.

| $n_1$ | 128 | 256 | 256 | 256 | 256 |
|---|---|---|---|---|---|
| $n_2$ | 128 | 128 | 256 | 512 | 1024 |
| RMSE | 3.26 / 3.58 | 3.22 / 3.51 | 3.20 / 3.49 | <u>3.17</u> / <u>3.46</u> | **3.15** / **3.42** |
| Runtime (s) | **0.17** / **0.01** | <u>0.20</u> / **0.01** | 0.22 / **0.01** | 0.27 / <u>0.02</u> | 0.36 / <u>0.02</u> |
| Number of parameter (M) | **1.1** / **0.6** | <u>1.7</u> / <u>1.1</u> | 2.3 / 1.7 | 3.5 / 2.9 | 5.8 / 5.9 |
| Model size (MB) | **4.5** / **2.8** | <u>6.7</u> / <u>4.5</u> | 9.0 / 7.3 | 14.0 / 13.0 | 23.0 / 24.2 |

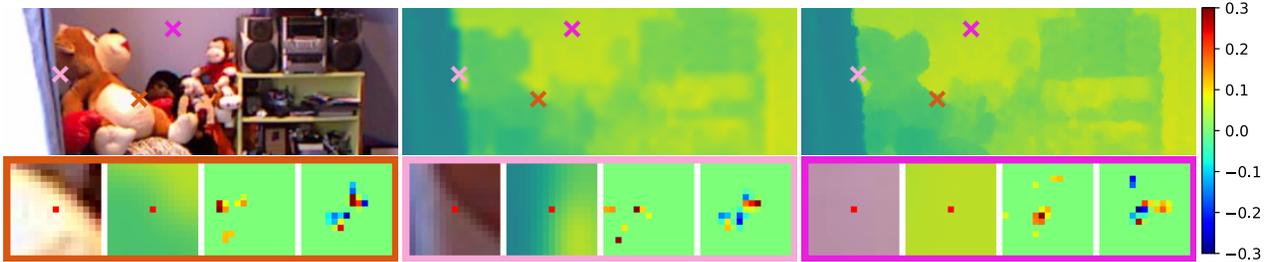

Figure 4. Visualization of filter kernels. Top: (From left to right) a RGB image, a low-resolution depth image, and an upsampling result by DKN. Bottom: (From left to right) Snippets of RGB images, low-resolution depth images, and kernels learned w/o and w/ the residual connection. The center positions in the RGB and depth images are denoted by red dots. The kernel weights are plotted with a heat map. (Best viewed in color.)

(*e.g.*, see Tables 1). The table shows that using more channels for feature extraction helps improve performance, but requires more runtime and a large number of parameters to be learned. For example, DKN takes twice more time for a (modest) 0.11 RMSE gain. Consequently, we choose the number of feature channels $n_1 = 128$ and $n_2 = 128$ for both models.

**Kernel visualization.** We show in Fig. 4 some examples of 3×3 filter kernels estimated by the DKN with/without the residual connection. Although the sampling positions are fractional, we plot them on a discrete regular grid using bilinear interpolation for the purpose of visualization. Corresponding kernel weights are also interpolated. We observe three things: (1) The learned kernels are spatially adaptive and edge-aware. For example, the kernels learned without the residual connection aggregate depth values that are similar to that at the center position. (2) They can handle the case when the structures from the HR color and LR depth images are not consistent as shown in the second example. (3) The kernels learned with the residual connection are orientation-selective and look like high-pass filters. For example, the kernels from the first and second examples can extract diagonal and vertical edges, respectively.

**Upscaling factors for training and testing.** Table 3 compares the average RMSE between DKN and DJFR [15] on the NYU dataset [21], when the scale factors for training and test are different. It shows that the performance is degraded for both methods in this case. This may be handled by a scale augmentation technique during training [12]. Another observation is that our model gives a similar result when the scale factor for testing is smaller than that used for training. For scale factors of ×4, ×8 and ×16, our model trained with factor of ×16 gives the average RMSE of 9.04, 8.61 and 6.51, respectively, whereas DJFR gives 19.12, 13.43 and 10.11, respectively. This demonstrates that our model generalizes better over different scale factors. A visual comparison is shown in Fig. 5.

Table 3. RMSE comparison (DKN/DJFR [15]) of using different upscaling factors for training and testing on depth map upsampling.

| Train/Test | 4× | 8× | 16× |
|---|---|---|---|
| 4× | **1.62**/ 3.34 | 6.70/10.21 | 11.24/19.75 |
| 8× | 3.93/ 9.27 | **3.26**/ 5.86 | 10.53/15.65 |
| 16× | 9.04/19.12 | 8.61/13.43 | **6.51**/10.11 |

Table 4. RMSE comparison by varying the number of training data on depth map upsampling (8×).

| Datasets | Methods | 10 | 50 | 100 | 200 | 500 | 700 | 1000 |
|---|---|---|---|---|---|---|---|---|
| NYU v2 [21] | DMSG [10] | 7.40 | 6.32 | 5.97 | 5.64 | 5.41 | 5.35 | 5.38 |
| | FDKN | 5.01 | 4.28 | 3.98 | 3.77 | 3.61 | 3.60 | 3.58 |
| | DKN | 4.73 | 3.97 | 3.62 | 3.33 | 3.27 | 3.25 | 3.26 |
| Sintel [1] | DMSG [10] | 8.62 | 8.08 | 7.65 | 7.46 | 7.27 | 7.21 | 7.24 |
| | FDKN | 6.04 | 5.37 | 5.16 | 5.05 | 4.98 | 4.98 | 4.96 |
| | DKN | 5.79 | 5.24 | 5.01 | 4.85 | 4.82 | 4.74 | 4.77 |

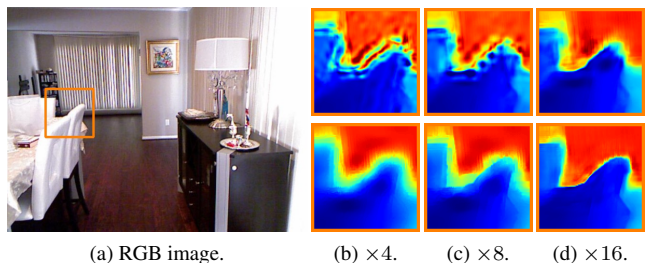

(a) RGB image.    (b) ×4.    (c) ×8.    (d) ×16.

Figure 5. Visual comparison of upsampled depth images for DJFR [15] (top) and DKN (bottom) when the scale factors for training (×16) and test (×4, ×8, ×16) are different.

**Kernel prediction vs. direct regression.** Our model has several advantages over current CNN-based approaches that directly regress the output. First, the direct regression may overfit the particular characteristics of training data, especially when the number of training samples is small. In contrast, weighted averaging smooths the output and acts as a regularizer, suggesting that our model is not seriously affected by the number of training samples. To demonstrate this, we evaluate the average RMSE performance in Table 4 when varying the size of the training data. We train the DKN, FDKN and DMSG [10], where the output is directly regressed from input images, for depth map upsampling (8×) while gradually increasing the number of training samples from 10 to 1, 000 in the NYU v2 dataset [21]. We test them in the same configuration as in Table 2 in the main paper. Table 4 shows that our models are more robust to the size of training data and generalize better to other images (*e.g.*, on the Sintel dataset [1]) outside the training dataset than the direct regression approach, even with more learnable parameters (1.1M for DKN and 0.6M for FDKN vs. 0.43M for DMSG [10]). In particular, the DKN trained with only 10 images outperforms the state of the art by a significant margin for all test datasets (see Table 2 in the main paper). Second, the kernels learned by direct regression are defined implicitly and hard to visualize. In contrast, our method learns sparse kernels (*i.e.*, *where* to aggregate) explicitly. We can interpret and visualize why kernels learned by our model give smooth results while preserving edges (Fig. 4), and this also gives a clue for tuning hyper-parameters. For examples, we can reduce the maximum range of offset locations (*i.e.*, the size of the filter kernel) and the number of weights (*i.e.*, the total number of samples to aggregate), when the weights are concentrated on central parts of the kernels, and a few of them are highly confident, respectively. Finally, our model can be applied to any tasks requiring an explicit weighted averaging processing beyond depth map upsampling, as confirmed for the task of semantic segmentation in Sec. 4.

## 3. More results

### 3.1. Depth map upsampling

We show in Fig. 6 more examples on depth upsampling (8×) on the NYU v2 [21], Lu [17], Middlebury [9] and Sintel [1] datasets.

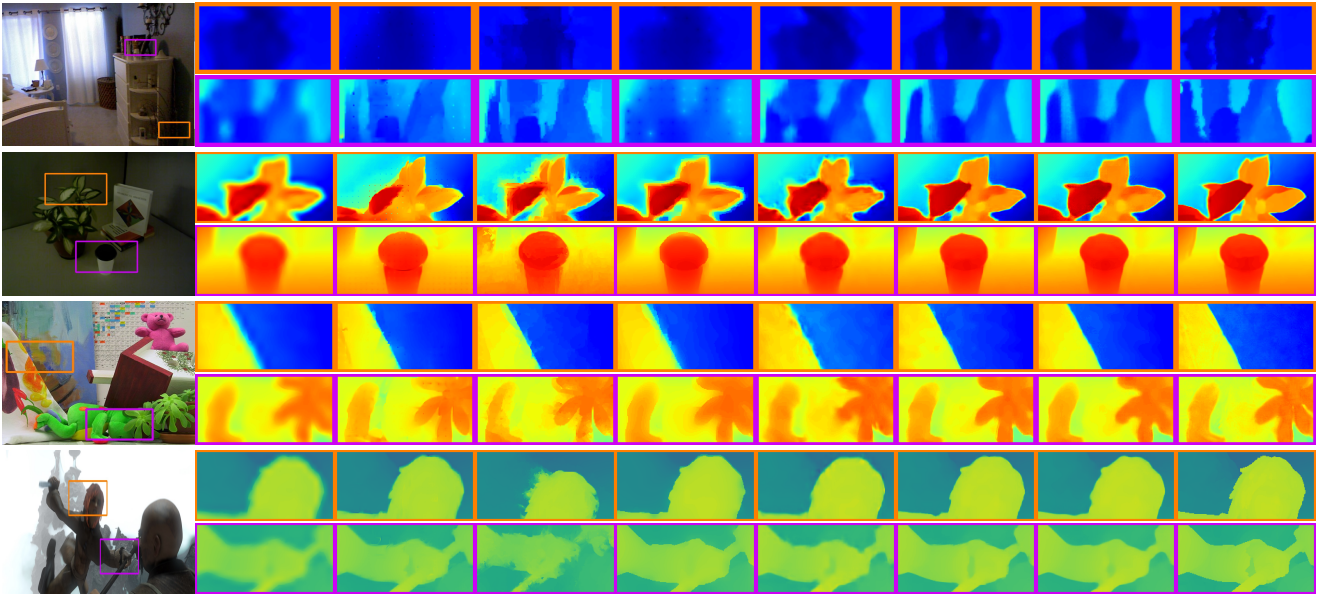

(a) RGB image.　(b) GF [7].　(c) TGV [4].　(d) Park [20].　(e) SDF [5].　(f) DJFR [15].　(g) DKN.　(h) FDKN.　(i) Ground truth.

Figure 6. Visual comparison of upsampled depth images (8×). Top to bottom: Each row shows upsampled images on the NYU v2 [21], Lu [17], Middlebury [9] and Sintel [1] datasets, respectively. Note that we train our models with the NYU v2 dataset, and do not fine-tune them to other datasets.

### 3.2. Saliency map upsampling.

To evaluate the generalization ability of our models on other tasks, we apply them trained with the NYU v2 dataset to saliency map upsampling without fine-tuning. We downsample saliency maps (×8) in the DUT-OMRON dataset [25], and then upsample them under the guidance of HR intensity images. We show in Table 5 a comparison of weighted F-measure [18] between upsampled images and the ground truth. Figure 7 shows examples of the upsampling results by the state of the art and our models. The results show that our models outperform others including a CNN-based method [15].

Table 5. Quantitative comparison on saliency map upsampling in terms of weighted F-scores [18]. We use 5,168 images from the DUT-OMRON dataset [25].

| Bicubic Int. | GF [7] | SDF [5] | DJFR [15] | DKN | FDKN |
|---|---|---|---|---|---|
| 0.8860 | 0.8895 | 0.8851 | 0.9248 | 0.9440 | **0.9606** |

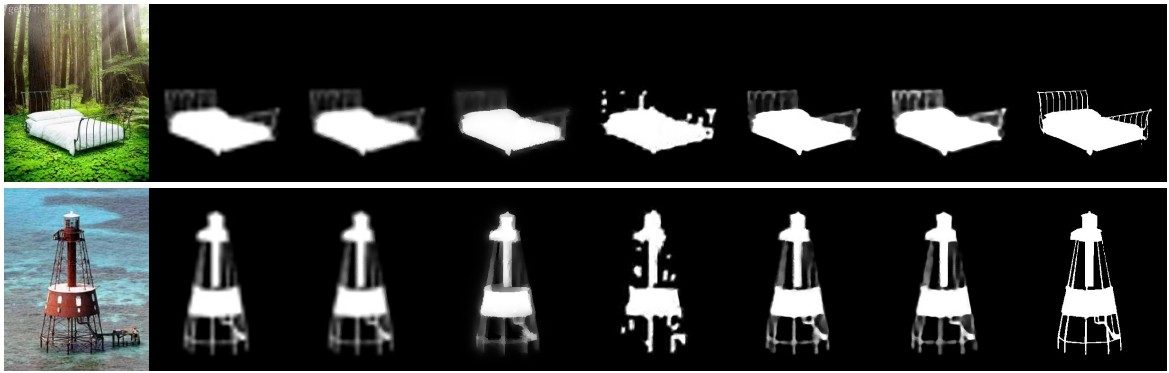

(a) RGB image.　(b) Bicubic Int.　(c) GF [7].　(d) SDF [5].　(e) DJFR [15].　(f) DKN.　(g) FDKN.　(h) Ground truth.

Figure 7. Visual comparison of saliency map upsampling (8×) on the DUT-OMRON dataset [25].

## 4. Other applications

In this section, we present other applications of our model including cross-modality image restoration, texture removal and semantic segmentation.

### 4.1. Joint noise removal

Following [14, 15], we use depth denoising as a proxy task for cross-modality image restoration and texture removal that require selectively transferring the structural details from the guidance image to the target one, since ground-truth datasets for these tasks are not available. For cross-modality image restoration and texture removal tasks, all previous works (e.g., [14, 24, 15, 7]) we are aware of for these tasks offer qualitative results only.

#### 4.1.1 Implementation details

**Training.** We train the networks for denoising depth images with RGB/D image pairs from the NYU v2 dataset [21]. The models for depth noise removal are similarly trained but with 4k iterations under the guidance of high-resolution RGB images. Noisy depth images are synthesized by adding Gaussian noise with zero mean and variance of $0.005$. For comparison, average RMSE for GF [7], Yan [24], SDF [5], DJF [14], and DKN on depth noise removal are 5.34, 12.53, 7.56, 2.63, and 2.46, respectively, in the test split of the NYU v2 dataset, showing that our model again outperforms the others.
**Testing.** We apply the models to the tasks of cross-modality image restoration and texture removal without fine-tuning. In case of a 1-channel guidance image (e.g., RGB/NIR image restoration), we create a 3-channel image by duplicating the single channel three times. For multi-channel target images (e.g., in texture removal), we apply our model separately in each channel and combine the outputs. We do not use the residual connection for noise removal tasks, since we empirically find that it does not help in this case.

#### 4.1.2 Results

**Cross-modality image restoration.** For flash/non-flash denoising, we set the flash and non-flash images as guidance and target ones, respectively. Similarly, we restore the color image guided by the flash NIR image in RGB/NIR denoising. Examples for flash/non-flash and RGB/NIR restoration are shown in Fig. 8. Qualitatively, our models outperform other state-of-the-art methods [14, 5, 7], and give comparable results to those of Yan [24] that is specially designed for this task. In particular, they preserve edges while smoothing noise without artifacts. This demonstrates that our models trained with RGB/D images can generalize well for others with different modalities.
**Texture removal.** We set the textured image itself for guidance and target, and apply our models repeatedly to remove small-scale textures. We show examples in Fig. 9. Compared to the state of the art, our models remove textures without artifacts while maintaining other high-frequency structures such as image boundaries and corners. A plausible explanation of why networks trained for denoising depth images work for texture removal is that textures can be considered as patterned

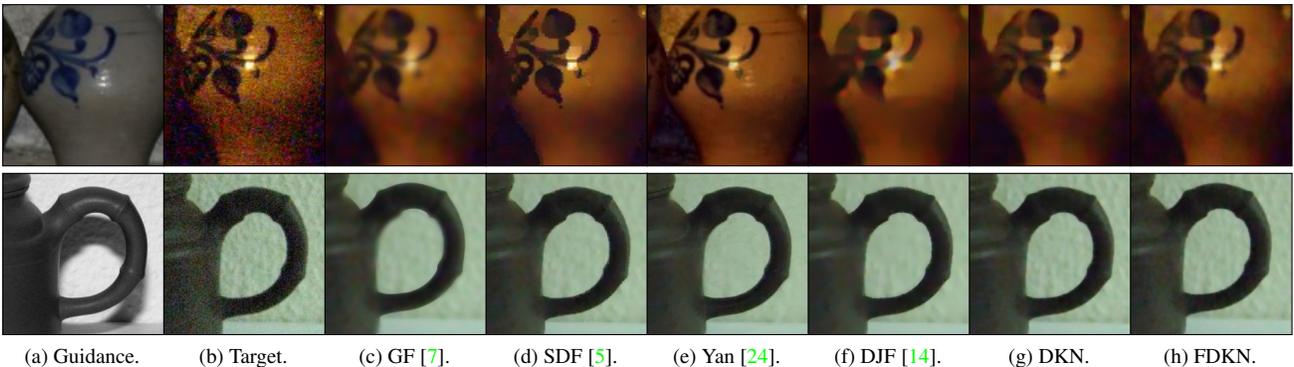

(a) Guidance.    (b) Target.    (c) GF [7].    (d) SDF [5].    (e) Yan [24].    (f) DJF [14].    (g) DKN.    (h) FDKN.

Figure 8. Examples of cross-modality noise reduction for (top) flash/non-flash denoising and (bottom) RGB/NIR denoising. Our models preserve textures while smoothing noise. GF and DJF tend to over-smooth the textures. Artifacts are clearly visible in the results of SDF. The method of [24], specially designed for this task, gives the best results.

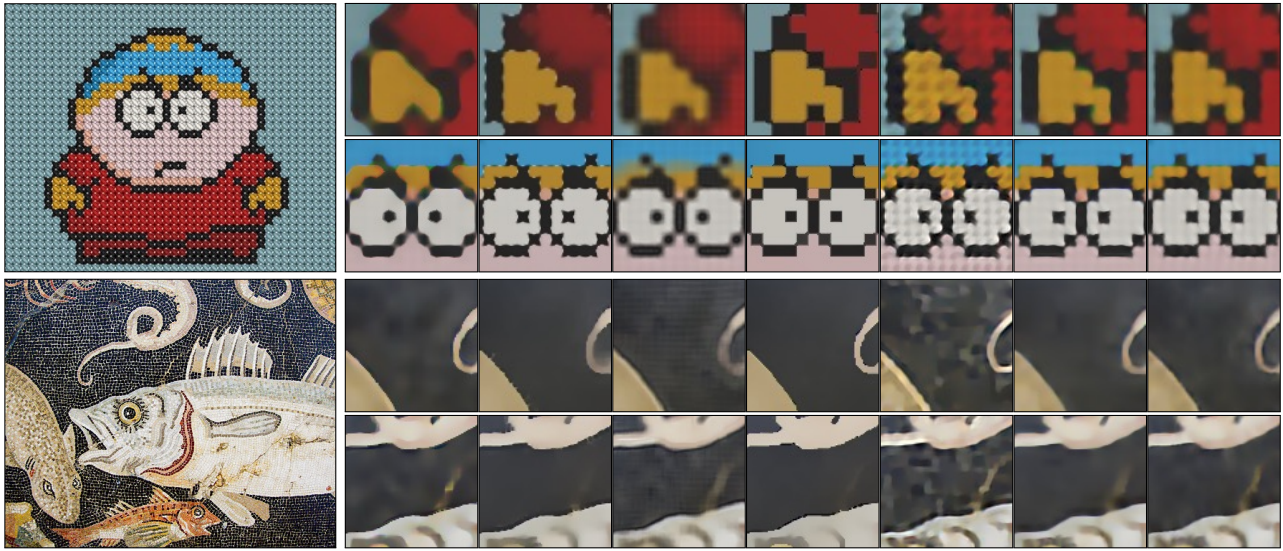

(a) Input image and snippets. (b) RGF [26]. (c) RTV [23]. (d) Cov [11]. (e) SDF [5]. (f) DJF [14]. (g) DKN. (h) FDKN.
Figure 9. Visual comparison of texture removal for regular (top) and irregular (bottom) textures.

noise. Repeatedly applying our networks thus removes them. A similar finding can be found in [14]. We empirically find that 4 iterations are enough to get satisfactory results, and use the same number of iterations for all experiments.

### 4.2. Semantic segmentation

CNNs commonly use max pooling and downsampling to achieve invariance, but this degrades localization accuracy especially at object boundaries [16]. DeepLab [2] overcomes this problem using probabilistic graphical models. It applies a fully connected CRF [13] to the response of the final layer of a CNN. Zheng *et al*. [27] interpret CRFs as recurrent neural networks which are then plugged in as a part of a CNN, making it possible to train the whole network end-to-end. Recently, Wu *et al*. [22] have proposed a layer to integrate guided filtering [7] into CNNs. Instead of using CRFs [2, 13, 27] or guided image filtering [7], we apply the FDKN to the response of the final layer of DeepLab v2 [2], before CRFs. Following the experimental protocol of [22], we plug FDKN in DeepLab-v2 [2], which uses ResNet-101 [8] pretrained for ImageNet classification, as a part of CNNs for semantic segmentation[1], instead of applying a fully connected conditional random field (CRF) [13] to refine segmentation results. That is, we integrate DeepLab-v2 and our model and train the whole network end-to-end, avoiding an offline post-processing using CRFs.

#### 4.2.1 Implementation details

**Training.** We use the Pascal VOC 2012 dataset [3] that contains $1,464$, $1,449$, and $1,456$ images for training, validation and test, respectively. Following [2, 22], we augment the training dataset by the annotations provided by [6], resulting in $10,582$ images, and use $1,449$ images in the validation set for evaluation. We train the network using a softmax log loss with a batch size of 1 for 20k iterations. The SGD optimizer with momentum of $0.9$ is used. As learning rate, we use the scheduling method of [2] with learning rate of $2.5 \times 10^{-4}$ and $2.5 \times 10^{-3}$ for DeepLab-v2 and FDKN, respectively.
**Testing.** We upsample 21-channel outputs (20 object classes and background) of DeepLab-v2 before a softmax layer using a high-resolution color image. We apply the FDKN separately in each channel.

#### 4.2.2 Results

We show in Table 6 mean intersection-over-union (IoU) scores for the validation set in the Pascal VOC 2012 benchmark [3]. To the baseline method (DeepLab v2 w/o CRF [2]), we add CRF [13], guided filtering [22], or FDKN layers. This table shows that our model quantitatively yields better accuracy in terms of mean IoU than other state-of-the-art methods. Examples of

---
[1] We use a PyTorch version available online: https://github.com/isht7/pytorch-deeplab-resnet

Table 6. Quantitative comparison on semantic segmentation in terms of average IoU. We use 1,449 images from the validation set in the Pascal VOC 2012 benchmark [3].

| Methods | Baseline [2] | DenseCRF [13] | DGF [22] | FDKN |
|---|---|---|---|---|
| Mean IoU | 70.69 | 71.98 | <u>72.96</u> | **73.60** |

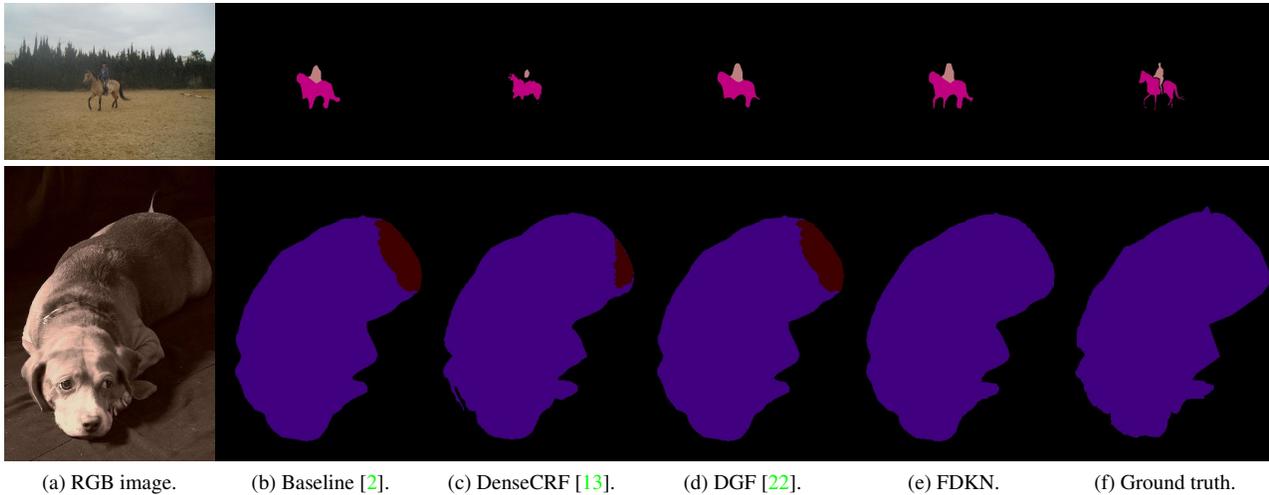

(a) RGB image.   (b) Baseline [2].   (c) DenseCRF [13].   (d) DGF [22].   (e) FDKN.   (f) Ground truth.

Figure 10. Visual comparison of semantic segmentation on the validation set of Pascal VOC 2012 benchmark [3]. Compared to the state of the art, our model shows better ability to improve the localization accuracy of object boundaries and refine incorrectly labeled segments.

semantic segmentation are shown in Fig. 10. Our model outperforms other methods qualitatively as well: It improves the localization of object boundaries (first row) and refines incorrect labels (second row).



\end{document}